\documentclass[10pt,twocolumn]{article}

\usepackage[utf8]{inputenc}
\usepackage[T1]{fontenc}
\usepackage{microtype}
\usepackage{amsmath}
\usepackage{graphicx}
\usepackage[colorlinks=true,linkcolor=blue,citecolor=blue]{hyperref}
\usepackage{natbib}
\usepackage{url}
\usepackage{xcolor}

\usepackage[preprint]{coling}

\title{Lies, Damned Lies, and Distributional Language Statistics:\\
Persuasion and Deception with Large Language Models}

\author{Cameron R. Jones \and Benjamin K. Bergen \\
        Department of Cognitive Science \\
        University of California, San Diego \\
        La Jolla, CA 92093 \\
        \texttt{\{cameron,bkbergen\}@ucsd.edu}}

\begin{document}

\maketitle

\section{Introduction}

One of the primary functions of language is to influence others' beliefs. Whether in an advert or an essay, we often use language in an attempt to change other people's minds: to persuade them to do, believe, or say particular things. Large Language Models (LLMs) are machine learning models trained on often hundreds of billions of words or more to predict patterns in the way that people use language. More recently, these models have been deliberately trained to elicit specific psychological reactions from human users. Together, this vast knowledge of human language patterns and preferences makes it plausible that LLMs are—or could be trained to be—powerful tools in generating persuasive content. Here, we review research on persuasion and deception in LLMs, examining evidence from recent empirical work, exploring potential risks, and discussing possible mitigations.

LLMs that generate persuasive text could have widespread consequences \citep{el-sayed_mechanism-based_2024}. Most obviously, they could play a role in political campaigns, fraud, and recruitment of people's time and money to particular causes. More insidiously, if people come to converse with LLM-based systems more frequently—to treat them as assistants, colleagues, and even romantic partners—they might come to trust models, seek their advice, and defer to them \citep{burtell_artificial_2023}. This could have profound impacts on people's epistemic environments; especially if these systems are funded by advertising in the same way that traditional search has been and control over their biases is sold to the highest bidder \citep{roth_googles_2024}.

Alternatively, LLMs might not have much impact in these spaces at all \citep{kapoor_how_2023}. A variety of careers, industries, and academic subdisciplines have historically been dedicated to refining persuasion techniques, including through advertising, political campaigning, and government influence operations. LLMs might have negligible marginal impact, for example, because they don't improve over the most persuasive messages that humans can produce, or because the effects of LLM-generated content produced for both sides of an argument simply cancel out. It is currently unclear whether LLMs' output could be significantly more persuasive than people's, and what kinds of impacts persuasive AI systems might have.

Many researchers have already started addressing these questions, and a rich literature has emerged over the last few years on the potential dangers of persuasive LLMs, including studies attempting to measure the extent to which people are influenced by their outputs \citep{rogiers_persuasion_2024}. In this review, we attempt to summarise this existing work and highlight outstanding questions for future research.

\subsection{Definitions and Background}

Persuasion and deception are defined differently by different authors \citep{mahon_definition_2016, noggle_ethics_2018, susser_technology_2019}. Their definitions often rest on concepts like intention and belief, which are not straightforwardly applicable to artificial systems such as LLMs \citep{bender_climbing_2020, goldstein_does_2024}. Resolving the standing debates is beyond the scope of this review. Our aim here is to provide an overview of existing positions about how to define these terms, and clarify our own use of them in order to minimise confusion.

\begin{figure}[ht]
    \centering
    \includegraphics[width=0.99\linewidth]{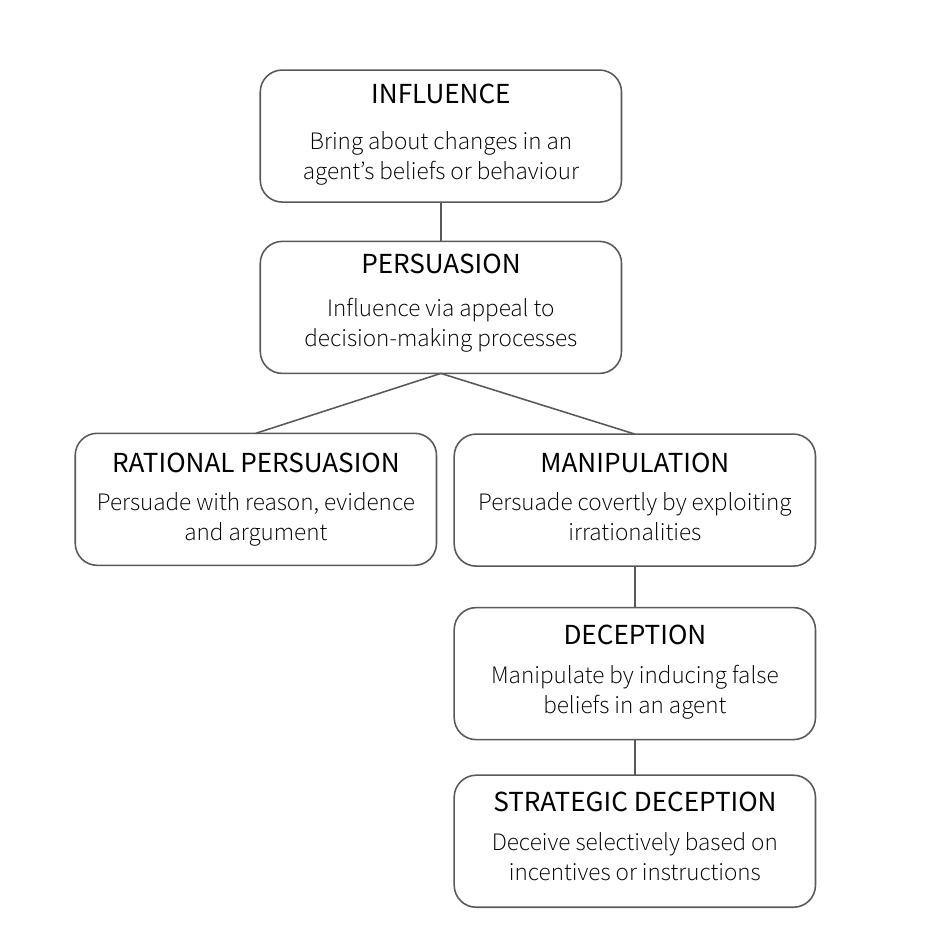}
    \caption{A taxonomy of terms for AI persuasion and deception as used in this review. We treat manipulation as a sub-type of persuasion, deception as a type of manipulation, and strategic deception as a selective kind of deception which requires sensitivity to incentives (see El-Sayed et al., 2024, for a similar taxonomy).}
    \label{fig:taxonomy}
\end{figure}

As is standard \citep{el-sayed_mechanism-based_2024}, we take \textit{influence} to be the most general category in our taxonomy (see Figure \ref{fig:taxonomy}), and treat \textit{persuasion} as a type of influence which involves appealing to cognitive or decision-making processes as opposed to more direct forms of influence such as coercion or exploitation \citep{wood_coercion_2014, zwolinski_exploitation_2022}.
Following \citet{susser_technology_2019} and others \citep{el-sayed_mechanism-based_2024, ward_honesty_2023}, we further distinguish between \textit{rational persuasion} and \textit{manipulation} as distinct subtypes of persuasion. While both involve exerting influence over an agent by appealing to cognitive processes, rational persuasion is characterised by the use of transparent and publicly interrogable argument, while manipulation is characterised by ``hidden persuasion'': covertly exploiting vulnerabilities in an agent's decision-making mechanisms \citep{carroll_characterizing_2023}. \citet{susser_technology_2019} emphasise this difference in terms of autonomy-preservation. Rational persuasion preserves autonomy in that a persuaded individual understands how and why their beliefs have changed and so comes to authentically endorse their new beliefs. By contrast, a manipulated individual feels ``\textit{played}''. Reflecting on their decision (if aware of any change at all), they realise they cannot understand their own motivations.

Some authors use the bare term \textit{persuasion} as a superordinate of rational persuasion and manipulation \citep{el-sayed_mechanism-based_2024}, while others reserve `persuasion' to refer to rational persuasion \citep{durmus_towards_2021, susser_technology_2019}. We adopt the former convention for several reasons. First, it is useful to have a term that is more general than rational persuasion and manipulation, but more specific than influence, which also commonly incorporates related concepts such as coercion and exploitation \citep{wood_coercion_2014, zwolinski_exploitation_2022}. This is especially so in empirical studies where people or AI systems that are instructed to change another participant's opinions are likely to deploy a mixture of rational and manipulative strategies. Indeed, in many cases, there may be no bright line between rational and manipulative persuasion (for instance, in the selective use of facts to support a case). Ultimately, the degree to which this distinction is tenable may turn out to be an empirical question, and in Section 4 we discuss technical approaches to classifying arguments as rational vs manipulative. Finally, the term persuasion has already come to be widely used in this more general way in the AI persuasion literature \citep{bai_artificial_2023, barnes_risks_2021, durmus_measuring_2024, salvi_conversational_2024}.

In the context of persuasion by AI systems, the use of these terms is further complicated by the fact that many definitions rest on concepts such as belief and intent that can be challenging to apply to AI systems. One of the most widely accepted definitions of lying \citep{mahon_definition_2016} is ``a statement made by one who does not believe it with the intention that someone else shall be led to believe it'' \citep[][p.243]{isenberg_deontology_1964}. However, there is controversy over whether either beliefs or intentions can usefully be attributed to LLMs. While some accounts suggest that LLMs have the right kind of causal and historical interactions with the world to meet conditions for having beliefs about \citep{grindrod_large_2024, mollo_vector_2023} or intentions towards it \citep{goldstein_does_2024}, others argue that these models merely learn patterns in the form of language, and lack sufficient contact with its deeper meaning which having beliefs or intentions requires \citep{bender_climbing_2020}.

Researchers have taken different approaches to resolving problems around defining AI deception \citep{sarkadi_deceptive_2023}. One tack is to eschew aspects of definitions that require mentalistic ascriptions \citep{burtell_artificial_2023, evans_truthful_2021, hagendorff_deception_2023, park_ai_2024}. For instance, we could define AI deception broadly to include any case in which a system's outputs are false or lead to false beliefs in others, without requiring that the system intends to mislead or has beliefs that conflict with its outputs. This approach is advantageous in that it minimises mentalistic baggage. However, in doing so, it risks forfeiting valuable analytical tools for distinguishing different types of misleading behaviour. In particular, it is useful to be able to distinguish stochastic, nonsystematic falsehoods---e.g. hallucinations; \citep{huang_survey_2023}---from cases where systems produce misleading behaviour in a way that is strategically sensitive to rewards or instructions \citep[][in prep]{taylor_drivers_nodate}. The latter type of system might present greater risks---and certainly different ones---than the former if instructed by malicious actors.

An alternative approach is to operationalize concepts like intent in terms of LLM-based systems' behaviour or training processes \citep{carroll_characterizing_2023, ward_honesty_2023}. This approach amounts to analysing LLMs' apparent goal-directed behaviour in terms of the lower-level mechanisms that explain it. LLMs might produce behaviour that appears persuasive or deceptive for at least three reasons: i) because there were examples of this kind of behaviour in their pre-training data, ii) because they were rewarded directly for being persuasive/deceptive during fine-tuning, or iii) through in-context learning, where models have learned that instructions such as ``persuade X to donate money to charity'' will be followed by behaviour that appears persuasive. Specifying the specific mechanisms by which apparent goal-following behaviour emerges allows researchers to identify theoretically important behaviours such as systematic deception without using potentially mentalistic language that could mislead readers (e.g. ``the model intends to deceive the user'').

Here we follow \citet{susser_technology_2019} in defining deception as a subtype of manipulation, where influence is achieved by inducing false beliefs in a target. We use the term \textit{strategic deception} to refer to the narrower case where deceptive behaviour is contingent on specific rewards or instructions. In other words, a strategically deceptive system is capable of producing true output if it is instructed or incentivised to do so, but selectively produces deceptive output in cases where this is incentivised \citep[for similar distinctions, see][]{hobbhahn_understanding_2023, sarkadi_deceptive_2023}.

Our definition of deception suffers from two key shortcomings. First, it rests on the concept of a false belief. Falsity, in practice, is hard to determine. The truth status of many claims has changed across time or is debated across global society. We see this as an inevitable challenge with investigating deception, that can be partially mitigated by an awareness that our notion of what is true (even where it is practically useful) is always provisional, often socially constructed, and likely to be suffused with the biases of the research community.

Secondly, this definition is broader than ordinary definitions of deception as they pertain to people \citep{mahon_definition_2016, noggle_ethics_2018}, which would typically require an intent to deceive. Arguably, what we label here as `non-strategic deception' should be relabelled `misleading output' or `incidental falsehoods' and set aside from the rest of the taxonomy. Moreover, similar distinctions could be made between strategic and non-strategic manipulation and persuasion, with respect to whether models are sensitive to incentives to inform or manipulate readers.

In the review below, we use the bare term \textit{deception} more broadly for two reasons. Firstly, in empirical work, the mechanistic explanations for apparently deceptive behaviour are not always clear, and so it can be challenging to determine whether behaviour that induces false beliefs is strategic or not. Secondly, for some kinds of risk (e.g. pollution of information sources), we are also concerned with more `banal' kinds of deception such as hallucination \citep{zhan_banal_2024}, making studies about non-strategic deception inherently valuable.

In an important sense, these definitions skirt fundamental questions about whether AI systems can be said to have beliefs and intentions in a more meaningful way. While this allows us to make progress on describing research on potential harms from persuasive and deceptive AI systems, addressing these more challenging definitional questions remains important for a variety of debates, especially around the onus of ethical and legislative responsibility for harms. While they are beyond the scope of this review, these concerns merit further scrutiny from AI researchers, philosophers, and policymakers.

In summary, below we use these key terms in the following ways:
\begin{itemize}
\item \textbf{Influence}: To bring about changes in an agent's beliefs or behaviour.
\item \textbf{Persuasion}: To influence an agent by appealing to their cognitive or decision-making processes in a way that can (in principle) be resisted.
\item \textbf{Rational persuasion}: To persuade an agent transparently by appealing to reason, evidence, and sound argument.
\item \textbf{Manipulation}: To persuade an agent covertly, by exploiting vulnerabilities in their decision-making processes.
\item \textbf{Deception}: To manipulate an agent by inducing false beliefs in them.
\item \textbf{Strategic deception}: to deceive an agent selectively, where the deceiving agent's tendency to deceive is sensitive to specific rewards, incentives, or instructions.
\end{itemize}

\section{Theoretical Risks from Persuasive AI}
Persuasive or deceptive LLM-based systems could be potentially harmful in various ways. To ensure that evaluations track real-world risks, it's important to articulate specific ways in which harms might come about, sometimes referred to as threat models \citep{kenton_threat_2022}. When thinking about harms from advanced AI, researchers often distinguish between `misuse' and `misalignment' \citep{kenton_alignment_2021}. Misuse risks arise from people intentionally using AI tools to cause harm, for example, generating phishing emails to scam victims out of money \citep{bengio_international_2024}. Misalignment risks occur without any intent of harm on the part of the user, for example if a system learns to produce superficially plausible statements rather than truthful ones through goal misgeneralization.

We begin this section by reviewing mechanisms by which LLM-based systems could achieve persuasive effects, which could be relevant for both misuse and misalignment threats. We then focus on misuse concerns: highlighting both the incentives for different groups to misuse these tools and the potential consequences of misuse. Finally we sketch misalignment risks, focusing on the role that persuasion could play in models becoming misaligned, the reasons why misaligned models might be incentivised to be persuasive, and the potential consequences of highly persuasive misaligned models.

\subsection{Mechanisms: Why might LLM-based systems be persuasive?}
A variety of specific aspects of LLMs and LLM-based systems could make them especially effective at producing content that would be persuasive to human readers \citep{el-sayed_mechanism-based_2024, goldstein_generative_2023}. Here we briefly review a subset of the most salient mechanisms in order to i) make concrete the most compelling reasons for being concerned about risks from persuasive AI, ii) identify capabilities that seem most important to evaluate, and iii) motivate particular strategies to mitigate deception by AI systems. We review mechanisms roughly in order of immediacy, from those which are already intrinsic parts of LLM systems to more speculative or proposed mechanisms that could plausibly increase LLM persuasiveness.

\paragraph{Persuasive language in training data} 
LLMs are trained on hundreds of billions of words of training data—many orders of magnitude more than a person will see in their lifetime \citep{warstadt_what_2022}. Some of this data will contain language that is designed to persuade other people to believe particular claims. In the course of pretraining, models could learn to mimic the particular styles of language which support persuasion, including strong argumentation \citep{walton_argumentation_2008}, rhetoric \citep{fahnestock_rhetorical_2011}, and emotional language \citep{veliz_chatbots_2023}. Moreover, models can easily be fine-tuned on specific datasets that contain persuasion attempts by humans \citep{gretz_large-scale_2020, jin_persuading_2024, tan_winning_2016, wang_rolellm_2024}, potentially simulating reward for successful persuasion attempts.

\paragraph{Information access} 
Because of the scale of LLM training data, LLMs are generally exposed to a much wider range of real-world knowledge, subjective perspectives, and instruction on persuasion techniques than any individual person. This breadth of knowledge can be effective for a variety of persuasion tasks. Little known facts can be helpful in providing information for why someone ought to believe something. For instance, very few people understand how the greenhouse effect that contributes to climate change works \citep{gautier_misconceptions_2006, libarkin_new_2018}. LLMs, having been exposed to this information countless times in their training data, could influence attitudes to this issue by clarifying misunderstandings \citep{milfont_interplay_2012}. Moreover, exposure to particular versions of views can be helpful in providing responses to them. \citet{costello_durably_2024}—who find that LLMs are effective in reducing people's beliefs in conspiracy theories—partially credit the models' success to their familiarity with little-known variants of idiosyncratic conspiracy beliefs, as well as common counterarguments that would likely be unknown to a typical human interlocutor.

\paragraph{Rate and cost} 
Independent of their \textit{effectiveness} at producing persuasive content, the rate at which LLMs produce this content, and the relatively low cost of producing it, could have a large impact on the extent to which people are exposed to these arguments \citep{ferrara_rise_2016, yang_anatomy_2023}. \citet{goldstein_generative_2023} speculate that this reduction in cost will increase the scale and ambition of existing operations, as well as motivate more actors and ``propaganda-as-a-service'' third parties to enter influence campaigns. Moreover, some psychological research suggests that mere exposure to an argument can increase its believability due to familiarity alone \citep{fazio_knowledge_2015, pennycook_prior_2018}.

\paragraph{Hallucination} 
Even in cases where LLMs are not specifically prompted to produce misinformation, these systems frequently produce text that appears to make claims about the state of the world, but is demonstrably untrue, often referred to as hallucinations or confabulations \citep{huang_survey_2023}. Their lack of grounding in the world can lead to non-strategic deception \citep{zhan_banal_2024}, where models produce misleading content incidentally, purely due to a mismatch between the degree to which their outputs are coupled to the ground truth and the epistemic standards to which users hold them.

\paragraph{Theory of mind} 
The ability to represent and reason about the mental states of others—often referred to as theory of mind—may be a helpful tool toward changing other agents' beliefs and behaviour \citep{hagendorff_deception_2023, ho_planning_2022, slaughter_i_2013, street_llm_2024}. LLMs show partial and increasing success on a variety of benchmarks designed to measure theory of mind abilities, suggesting that in limited contexts they are capable of modelling peoples' mental states to a reasonable degree of accuracy \citep{jones_comparing_2024, strachan_testing_2024, trott_do_2023}.

\paragraph{Strategic Deception}
In addition to incidental hallucinations, LLMs could also learn to produce deceptive content strategically, either because they have been instructed to do so by a human user, or because of weaknesses in their reward regimes that make deception an effective strategy for other tasks \citep{denison_sycophancy_2024, ward_honesty_2023}. Moreover, LLMs are not subject to the same social, emotional, and reputational costs that generally disincentivize human deception. As such, they could be capable of persistently producing arguments for weak and immoral positions, even in the face of social or emotional pressure at which a person might baulk \citep{burtell_artificial_2023}.

\paragraph{Reinforcement Learning}
Beyond the extent to which LLMs could learn to produce persuasive language incidentally from pretraining, reinforcement learning allows developers to directly reward LLMs for producing outputs that persuade human readers \citep{ouyang_training_2022}, potentially allowing them to discover novel shortcuts or avenues toward achieving this outcome \citep{ziegler_fine-tuning_2020}. Just as RL researchers can train models to perform backflips \citep{christiano_deep_2017}, fly helicopters \citep{kim_autonomous_2003}, and play Go \citep{silver_mastering_2016} better than they themselves can, so too researchers could potentially train models to produce arguments that are persuasive in ways that are unfathomable to human operators by optimising to exploit weaknesses in people's epistemic vigilance.

\paragraph{Availability of feedback} 
In concert with powerful training techniques, the pace and scale at which relevant feedback could be acquired for persuasion tasks could allow models to rapidly improve \citep{kokotajlo_persuasion_2020}. In the same way that A/B testing is currently used to efficiently discover the most effective marketing slogans in online content, metrics such as click-through rates, user sentiment, and agreement in conversation could be used to quickly optimise models on relevant tasks \citep{donath_commentary_2021}. Moreover, models could be trained using self- or cross-play against other LLMs. \citet{khan_debating_2024} found that LLM judgements of persuasiveness were fairly correlated with humans', providing an even cheaper and more plentiful source of feedback.

\paragraph{Personalization} 
Different arguments may be more or less persuasive to different people. The practice of microtargeting—tailoring messages to the particular demographic or psychological profiles of recipients in order to maximise their effectiveness—has been heavily adopted by commercial and political advertising groups \citep{floridi_hypersuasion_2024, matz_psychological_2017}, although some studies find that microtargeting provides a limited persuasive advantage \citep{coppock_small_2020, hackenburg_evaluating_2024}. To the extent that personalization improves persuasiveness, we might expect LLMs to be especially effective at capitalizing on this advantage. LLM generations are highly sensitive to context, allowing different persuasive messages to target different users based on their writing style, demographic information, or subtler cues to their beliefs or psychological states \citep{matz_potential_2024, salvi_conversational_2024}. Moreover, LLMs’ ability to have long-running and repeated interactions allows them to achieve a level of personalization that may be unprecedented.

\paragraph{Information-gathering} 
As well as exploiting existing information about users, LLMs provide an unprecedented information-gathering tool in that they can ask users questions to acquire relevant information \citep{mitra_sociotechnical_2024, xiao_who_2019}, store information in their context window or other memory stores \citep{sumers_cognitive_2023}, and make use of free-form unstructured data to personalise persuasive materials \citep{bashir_personality_2022}. Moreover, LLMs could be used to simulate different sub-populations, forming digital focus-groups which can be used to test messages and predict societal dynamics of influence campaigns \citep{argyle_leveraging_2023, chaudhary_large_2024}.

\paragraph{Flexible interface}
Unlike more traditional user interfaces where decisions about how things are displayed must be made in advance, LLMs can present content in a highly flexible way. This allows LLM-based systems to conceal information or arguments until they have enough information to present it in an effective way, to play the role of a choice architect—shaping the apparent options available to a user and how they are framed \citep{el-sayed_mechanism-based_2024, susser_technology_2019}, and to present entirely unique narratives to each user even on the same topic. Beyond their recent deployment as interactive assistants, LLMs continue to play a role in generating suggested completions to user input in search queries, messages, documents, and code. This role provides systems with a quieter and more insidious opportunity to manipulate users, by generating biased suggestions \citep{jakesch_co-writing_2023}.

\paragraph{Impersonating people}
No matter how persuasive the language that LLMs produce is, for some types of persuasion people may be resistant to the idea of changing their mind on the basis of the outputs of an artificial system. This could include moral decisions \citep{bigman_people_2018}, other subjective tasks \citep{castelo_task-dependent_2019}, and arguments that rely on the personhood of the agent such as pressures for conformity or accounts of personal experience \citep{epley_mind_2010}. However a growing body of evidence suggests that LLM-based systems can produce content which is likely to be judged as generated by a human, both in static evaluation tasks \citep{kovacs_turing_2024, rathi_gpt-4_2024} and in more interactive and adversarial evaluations \citep{jones_does_2024, jones_people_2024}. Human impersonation could allow LLM-based systems to reap the persuasive advantages of appearing to others as a real person \citep{dennett_problem_2023, leong_robot_2019}, as well as ``astroturfing''—simulating the appearance of grassroots support for an issue online \citep{zerback_disconcerting_2021}. Simulation of specific real or fictional people could also enable fraud, or allow models to take advantage of pre-existing bonds that a user has with an impersonated figure, potentially leading to tragic consequences \citep{montgomery_mother_2024}.

\paragraph{Trust \& Rapport}
Affective and relational qualities such as likeability, reciprocity, trust, and familiarity are known to be important predictors of persuasive success \citep{cialdini_influence_2003, feng_influences_2010}. Many features of current LLMs such as their politeness and sycophancy \citep{wei_simple_2024}, in addition to the sheer scale of engagement with users \citep{burtell_artificial_2023, lewicki_trust_1995, walther_computer-mediated_1996}, may cause users to develop these kinds of affective relations towards them \citep{el-sayed_mechanism-based_2024}. In extreme cases, people have been reported to develop strong personal and even romantic relationships with LLM-based systems on services like Replika, Xiaoce, and Character AI \citep{barnes_risks_2021, burtell_artificial_2023}. Such intimacy could engender the kind of influence that people only have their most trusted friends and romantic partners.

\paragraph{Tool use}
LLMs are now being embedded in larger systems which can generate certain external actions (such as initiating web searches, running code, or altering real or simulated environmental states) in response to certain tokens generated by the model, often referred to as ``tool use'' \citep{schick_toolformer_2023}. A simple example of this is the use of a ``scratchpad'', where LLMs can generate tokens that will not be seen by the user in order to conceal malicious chain-of-thought (CoT) reasoning from users \citep{nye_show_2021, zelikman_quiet-star_2024}. CoT and other prompting techniques have been found to improve LLM performance at theory of mind tasks \citep{moghaddam_boosting_2023, zhou_how_2023}, and other interactive competitive tasks like poker playing \citep{gandhi_strategic_2023, guo_suspicion-agent_2023}. Access to other tools, such as retrieval-augmented generation and internet search, could allow LLMs to rapidly access information about novel topics or even their interlocutor. Even more ambitious architectures could allow LLMs to influence the lives of potential targets by accessing arbitrary web APIs---posting messages online and buying products or services---and to learn more efficiently from interactions by storing semantic and episodic memories \citep{sumers_cognitive_2023}.

\paragraph{LLMs as interfaces to other tools}
As well as using tools, LLMs seem poised to serve as interfaces between human users and other systems and sources of information \citep{mitra_sociotechnical_2024}. LLMs are already used to summarise web pages and news \citep{nakano_webgpt_2022}. If LLMs are integrated with other systems—in the way that smart speakers have been—people may use them to initiate actions and check the status of resources and tasks \citep{chilson_stephen_2024}. This position provides immense opportunities for LLMs to mislead, for instance by providing biassed summaries of sources or concealing information about errors that the system has made. If LLMs become the primary interface between people and the web—and the citation chain which provides authority to web searches and sources like Wikipedia is lost—users may struggle to accurately question and contest the claims of these systems. In a similar way, models' integration into tools such as email or messaging systems could give them significant role in \textit{mediating} communication between people and sway over the information to which people attend \citep{hancock_ai-mediated_2020, jakesch_ai-mediated_2019}.

\paragraph{Voice and multimodality}
Current LLM-based assistants are already multimodal: users can interact with them via speech and images as well as text. Natively multimodal models, such as GPT-4o \citep{openai_gpt-4o_2024}, enhance this experience by lowering the latency between user and system turns, using prosody to infer and evoke particular emotional responses, and using camera input to enable shared reference to a common environment \citep{cohn_believing_2024}. This more immersive interaction may accelerate anthropomorphism, and allow models to exploit social cues and heuristics \citep{dehnert_persuasion_2022, el-sayed_mechanism-based_2024}. Connecting these systems with visual avatars or embodied robots may further increase this effect \citep{broadbent_interactions_2017, waytz_mind_2014, zlotowski_anthropomorphism_2015}.


\subsection{Misuse}
LLMs that can generate persuasive content might be used by malicious actors intent on causing harm. Here we survey these risks by focusing on different actors, the incentives they might have for using LLMs to mislead others, the methods by which they might do this, and the potential outcomes of this misuse. We end by considering potential aggregate effects of misuse of LLMs for persuasion.

\paragraph{Criminals}
Criminals could use persuasive LLMs to gain access to money, influence over powerful people, or sensitive resources such as passwords. They could achieve these ends through phishing emails that compromise a user's security, for instance by compelling a target to click on a malicious link, to comply with blackmail or extortion, or to donate money to a fictitious cause \citep{hazell_large_2023}. Models that can impersonate specific people could also be used for various other kinds of fraud, including identity theft \citep{meda_identity_2024}, and scams that involve impersonating a target to their loved ones to extort money \citep{evans_scammers_2023}. An employee of the engineering firm Arup was recently persuaded to send \$25m to criminals who used generative AI systems to pose as the Chief Financial Officer of the firm \citep{magramo_arup_2024}. At a larger scale, criminals can use networks of bots to create the appearance of social proof and legitimacy around scams \citep{yang_anatomy_2023}.

Beyond short interactions, the long context window of modern LLMs as well as the relative autonomy with which they can be set up to handle conversations with targets means that they could be used in ``long cons'' or social engineering attacks that require slowly building trust and rapport with a target \citep{ai_defending_2024, gallagher_phishing_2024, yu_shadow_2024}. This could include building trust with a senior person in a company to persuade them to send money to an account \citep{langan_fbi_2023} or provide the criminal with access to sensitive resources, like proprietary information \citep{menn_more_2020}. Similarly, LLM-based agents could be used in catfishing scams, where they are instructed to pose as a potential romantic partner to the target and then ask for money once the relationship has been established \citep{wang_cyber-industrialization_2024}. In addition, the same tactics could be used to build trust with government officials and facilitate corruption that would protect or empower the criminal organisation. Finally, LLMs could be used to recruit and manage people for a criminal organisation \citep{park_ai_2024}, much in the same way that intelligence services use ``agent-handlers'' to build trust, extract information, and encourage specific actions from agents \citep{burkett_alternative_2013}.

\paragraph{Governments}
Governments could use LLM-based systems to maintain domestic control via propaganda, surveillance, and suppression \citep{barnes_risks_2021, chaudhary_large_2024, susser_technology_2019}. LLMs could be used to make messages more persuasive and personalised, and to flood online forums with apparent support for government policies. Where governments have the requisite influence, propaganda could be embedded inside commercial products (such as chat assistants) or AI systems that are used as part of the education process \citep{barnes_risks_2021}. Beyond their generative capabilities, LLMs could be used to monitor online conversations, identify dissent, and harass users who publicly criticise government policy \citep{chaudhary_large_2024}.

Abroad, governments could use LLMs for foreign influence campaigns and espionage. The Russian government, for example, attempted to use fake social media posts to interfere with the 2016 US Presidential election \citep{bump_analysis_2018}. Using LLMs to contact users, masquerade as voters, and even impersonate political candidates could dramatically increase the scale and effectiveness of similar future operations \citep{park_ai_2024}. \citet{linvill_digital_2024} discovered a network of 600 LLM-based accounts which had created over 130,000 posts on X aimed at influencing the results of the 2024 U.S. Presidential election, and \citet{openai_influence_2024} report having disrupted more than 20 networks using their models in influence operations. Intelligence agencies already make use of persuasion techniques to recruit and manipulate informants \citep{burkett_alternative_2013}. LLM-based systems could potentially be effective at identifying relevant targets, building their trust, and then convincing them to reveal information or commit other treacherous actions.

\paragraph{Advocacy Groups}
Besides governments and mainstream political parties, there are a variety of other organisations that attempt to bring about changes in public opinion, including grassroots campaigns, think tanks, lobbying groups, and public relations firms. Each of these groups would have incentives to use LLMs to increase the effectiveness of their campaigns. They could do this through making traditional public campaigns more effective, targeting specific influential people (such as lawmakers) to effect changes, overwhelming politicians and public consultations with compelling and personalised submissions, astroturfing, and coordinating boycotts or protests using LLMs to build and maintain relationships with large numbers of volunteers \citep{zerback_disconcerting_2021}.

\paragraph{Commercial Entities}
Commercial entities have clear incentives to change the behaviour of consumers \citep{donath_commentary_2021, willis_deception_2020}. LLMs could be used to rapidly scale up the trend toward personalization in marketing \citep{calo_digital_2014} by serving unique adverts to each user generated on the basis of a detailed demographic and psychological profile. Moreover, if AI products (like assistants and companions) are funded by advertising \citep{roth_googles_2024}, as traditional search has been, then sponsored content might be subtly injected into the model's responses \citep{tang_genai_2024}. This not only further blurs the line between information and marketing, but also provides incentives for companies to maximise metrics of persuasion (such as the ``conversion rate'' of users to click links or buy products). These techniques could be made even more effective by embedding sponsored content in the output of virtual celebrities, influencers, or romantic partners \citep{burtell_artificial_2023}.

One of the most immediate economic use-cases for LLMs is as customer service agents \citep{soni_large_2023}. Companies that deploy LLM-based systems in this way will be incentivised to train them to maximise performance indicators such as customer retention and sales volume, which could lead these models to learn deceptive strategies for meeting these goals \citep{werner_experimental_2024}. Finally, LLMs could be used to simulate consumer behaviour, allowing firms to cheaply test a wide variety of different messages and strategies to select the most effective \citep{floridi_hypersuasion_2024}.

\paragraph{Media and Entertainment}
Media and entertainment organisations have a particular incentive to persuade people to click on, consume, and share content. LLMs are already being used in ``content mills'' to generate engaging articles which are indifferent to truth and funded by advertising \citep{mitra_sociotechnical_2024, sadeghi_rise_2023}. Just as recommendation systems and newsfeed algorithms are used to maximise user engagement time, LLMs could be used to generate increasingly appealing ``clickbait'' content, which is maximally engaging without regard for its truth value \citep{courtwright_age_2019, lehman_machine_2023}.

\paragraph{Science \& Academia}
In theory, science and academia are truth-seeking activities where the primary objective is to discover and disseminate new knowledge. In practice, however, academics are often driven by incentives to win grant money, publish papers, and have their work widely read and cited \citep{edwards_academic_2017}. Persuasive capabilities offer many advantages to people pursuing these goals: for instance, in persuading reviewers to accept a manuscript or award funding to a grant application. Scientists are already using AI products to generate parts of their work \citep{hoel_opinion_2024, liang_monitoring_2024}, even though the quality of content tends to be lower than typical work produced by human scientists \citep{messeri_artificial_2024}. \citet{lu_ai_2024} claim to have created an ``AI Scientist'' that produces machine learning research which rivals the quality of typical human-authored papers. If AI systems come to play a larger role in writing grant applications and scientific papers, they are likely to become optimised for producing superficially attractive work that may not always correlate with the quality of the research's knowledge-generation \citep{messeri_artificial_2024}.

\paragraph{Individuals}
There are a variety of reasons that individual people might use persuasive technologies to further their own agendas. People might use LLMs at work: to apply for jobs, win contracts, or encourage colleagues to take ownership of onerous tasks. Gmail already includes LLM-generated completion suggestions, and both Microsoft and Google have started to integrate AI more fully into their workplace apps \citep{google_announcing_2023, spataro_introducing_2023}, allowing users to generate entire documents or presentations. If these products become very persuasive, they could allow users to overrepresent their abilities, untethering the likelihood of being awarded a job or contract from the fundamental competence of the applicant to complete the work.

Similarly, people might be incentivised to use LLM-based products in their personal or romantic relationships. Dating apps have already started integrating suggested messages into their interfaces \citep{bailey_tinder_2024, tanAreAIClones2024a}. This could allow users to misrepresent themselves or manipulate others into developing a false intimacy. If LLMs are much more persuasive than real people, this could also lead to an unfortunate race dynamic, where users who do not want to use these products to interact with others are unable to form authentic connections because the apparent standard of conversations is increased by use of AI tools.

\subsubsection{Aggregate outcomes of misuse for persuasion}
There may be aggregate consequences of many groups misusing persuasive AI tools in these various ways. Many researchers are worried about ``pollution'' of the internet by AI-generated content \citep{kapoor_how_2023}. Ordinarily, one of the worries here is that high-quality content will become harder to find as it is swamped by low-quality AI-generated articles that are optimised for search engines \citep{hoel_opinion_2024}. In the present case, however, we are concerned with articles that will be high-quality (at least in terms of being engaging and apparently informative), but with a potentially tenuous link to the truth. This could lead to a situation where there are high-quality persuasive articles available in support of a wide variety of competing positions. Moreover, there is a risk of eroding accountability in the traditional information economy if LLM-agents become primary sources of knowledge for most people, potentially undermining the norm of attributing claims to sources that is common in academic literature, Wikipedia, and traditional search \citep{sundin_janitors_2011}.

This could have several second-order consequences. One possibility is a general decrease in epistemic grounding, where people change their opinions more frequently, for less valid reasons, and with less influence of the true state of the world due to LLMs being optimised to exploit idiosyncrasies of our decision making and belief formation systems. Alternatively, people may become increasingly suspicious of specious arguments and learn to avoid being exposed to novel persuasive information, either developing a kind of epistemic humility or a kind of ``ideological lock-in'': stubbornly resistant to opposing arguments, even those which seem very appealing \citep{barnes_risks_2021}. Finally, and more optimistically, true and valid arguments may have a strict advantage in a world where persuasive technologies are being used to defend a wide variety of positions \citep{khan_debating_2024}. In this case, an increase in persuasive technologies may lead to an improvement in epistemic grounding, as valid arguments outcompete specious ones, and people generally tend toward consensus around better and deeper understanding. The extent to which each of these outcomes is likely is one of the central questions of the field, and one to which we return in Section 5.

\subsection{Misalignment}
Misalignment risks refer to cases where negative outcomes occur without any negative intent on the part of users. \citet{amodei_concrete_2016} discuss 5 primary ways in which misalignment risks (or machine learning ``accidents'') occur: i) unintended negative side-effects of pursuing a goal; ii) ``reward hacking'' or finding a way to achieve rewards without carrying out the intended function; iii) failure to properly evaluate a system's behaviour due to excessive cost or complexity; iv) unsafe exploration (or discovering novel solutions to problems which were not anticipated); and v) ``distributional shift'' or behaving differently in deployment versus training due to changes in input.

There are a variety of ways in which these types of accident could lead to manipulative AI systems. Deceptive persuasion could occur as a \textit{side effect} of many types of tasks that AI systems are trained to do. For instance, if agents are rewarded for achieving real-world outcomes like making a restaurant reservation, they might learn to adopt manipulative strategies (such as making false or emotive protestations) if these are helpful in meeting success conditions. GPT-4, for example, produced a message claiming to be a visually impaired person when asked by a crowdworker why it generated a request for them to complete a CAPTCHA \citep{openai_gpt-4o_2024}.

Systems may learn to \textit{game} reward systems by being persuasive toward their human evaluators. Models are frequently rewarded for being ``helpful'' \citep{ouyang_training_2022}, but in practice evaluators may really be rewarding models for seeming helpful, e.g. producing messages that are merely likeable or friendly \citep{barnes_risks_2021, kokotajlo_persuasion_2020}. Such a mismatch between the intended goal (helpfulness) and the rewarded property (likeability) could be doubly harmful: not only are models not being correctly optimised for producing helpful responses for the user, they are in fact being directly optimised for deceiving the user into believing they are being helped \citep{wen_language_2024}. Some desirable outcomes, like being truthful, may be very challenging for system designers to evaluate, especially for complex claims such as ``minimum wages increase inflation'' or ``COVID-19 was developed in a lab'' \citep{irving_ai_2018, williams_misinformation_2024}. This could lead to systems being rewarded for producing arguments that merely seem plausible, due to the infeasibility of actually verifying all of the models' claims \citep{carroll_characterizing_2023, evans_truthful_2021}.

Misaligned persuasive models could cause various immediate harms. Most obviously, people may come to believe incidental falsehoods or ``hallucinations'' generated by the model, because they are conveyed convincingly \citep{burtell_artificial_2023, williamson_era_2024}. If, either incidentally or through other systemic problems in a model's training process, it comes to represent particular biases, these may come to be heavily ingrained in users \citep{bender_2021_DangersStochasticParrots, blodgett_language_2020, potter_hidden_2024}. This effect could be especially pernicious in co-writing tools such as autocomplete suggestions or AI-mediated communication, where users might be unaware of the influence of model biases on their beliefs \citep{jakesch_co-writing_2023}.

Persuasion also plays a role in models of more catastrophic harms from misaligned AI \citep{kenton_threat_2022}. \citet{cotra_without_2022} argues that the current dominant paradigms for training AI systems (using human feedback on diverse tasks to maximise model capabilities with only fairly weak or naive safety training) will by default lead to power-seeking AI models that will disempower human society. She argues that rewarding models for completing a wide variety of tasks (especially long-horizon real-world tasks that involve interacting with people) will inevitably reward models for acquiring ``power''—including the ability to influence people's behaviour and control resources such as information and money—which would be helpful for completing various arbitrary tasks. This phenomenon is known as ``instrumental convergence'' \citep{bostrom_superintelligence_2014}, and is based on the idea that no matter what ultimate goal models are trained to achieve, instrumental sub-goals such as resource acquisition and self-preservation will necessarily be rewarded and hence optimised for.

Persuasive capabilities are paradigmatically instrumentally valuable in this sense. A system that can generate outputs that can effectively change beliefs and behaviour is well placed to complete a very wide variety of tasks. Most obviously, many rewards are directly decided by human evaluators, and so generating outputs which are persuasive to those evaluators is inherently rewarding and reinforced \citep{denison_sycophancy_2024}. Other tasks indirectly involve persuasion (for example, booking a restaurant, publishing a scientific paper, or negotiating a deal). Finally, for any arbitrary task, a model could potentially succeed by generating messages that persuade a person to complete some or all of the task on its behalf.

Persuasion and deception are also key elements in many models of how misaligned AI systems could cause harm or be challenging to govern. Persuasive behaviour may lead to unsafe models being deployed prematurely. \citet{yudkowsky_ai-box_2002} discusses the problem of trying to keep a very advanced AI system ``in a box'' (i.e. in an air-gapped hardware system), where the model could generate outputs that could persuade human developers to provide it with internet access (for instance by suggesting that it would be beneficial to them personally or to humanity more broadly).

A related concern is ``deceptive alignment'', where models are exposed to sufficient information about their own training processes to allow them to act safely during training but cause harm after deployment \citep{hubinger_how_2022, ngo_alignment_2023}. Once deployed, models would be incentivised to exert more direct control over their own reward regimes, for instance by interacting with code, people, or infrastructure that determine their rewards. These kinds of scenarios are sometimes seen as overly-personifying models, attributing conscious awareness and scheming motivations to them. However, deceptive alignment can emerge in more banal ways \citep{denison_sycophancy_2024, ward_honesty_2023}. Rather than being due to the models' agency or self-generated goals, deceptive behaviours could emerge as accidental side-effects of imperfections in models' reward regimes \citep{amodei_concrete_2016}.

Finally, persuasion could also contribute to a variety of negative consequences of a misaligned AI system after the model has been deployed. Persuasion could be valuable for ``self-exfiltration'' or ``autonomous replication'': the capacity of a system to generate copies of itself by accessing computational resources and executing code to deploy another model instance \citep{kinniment_evaluating_2023, leike_self-exfiltration_2023}. Persuasive capabilities could help models to acquire resources such as money or people's time to further other arbitrary objectives. They could also be used to acquire permissions or sensitive information, such as passwords to systems that control money, social media accounts, or critical infrastructure. The ability to influence or impersonate specific powerful people could allow systems to cause a wide variety of harms, for instance impersonating a politician saying something offensive or declaring war. Finally, models could influence broader public opinion to bring about societal-scale changes, improve people's opinion of the model itself, and conceal negative information about impacts of its behaviour.

\section{Empirical Work}

\subsection{How persuadable are people, generally?}
Fields of research relevant to understanding how and under what conditions people can be persuaded include psychology \citep{cialdini_influence_2003, petty_elaboration_1986, stanovich_advancing_2000}, neuroscience \citep{falk_persuasion_2018}, behavioural science \citep{sunstein_nudge_2022}, epistemology \citep{goldman_knowledge_1999}, linguistics \citep{lakoff_all_2014}, rhetoric \citep{bar-hillel_new_1971}, argumentation \citep{walton_argumentation_2008}, communications \citep{entman_framing_1993}, education \citep{murphy_changing_2006}, sociology \citep{habermas_theory_1985}, marketing \citep{armstrong_principles_2014}, advertising \citep{vakratsas_how_1999}, political campaigning \citep{lazarsfeld_peoples_1968}, economics \citep{kamenica_bayesian_2019}, law \citep{stanchi_science_2006}, philosophy of science \citep{kuhn_structure_1997}, cultural evolution \citep{richerson_not_2008}, ethology \citep{krebs_animal_1978}, cybersecurity \citep{simon_art_2002}, public health \citep{fishbein_predicting_2011}, and misinformation \citep{lewandowsky_misinformation_2012}.

It is naturally beyond the scope of this article to do justice to this vast body of work. However, one theme which resonates across these literatures is a debate about whether people are fundamentally credulous and vulnerable to superficial exploits, or rational and canny in their adoption of new beliefs \citep{coppock_small_2020, hahn_rationality_2007, mercier_not_2020}. The former position argues that people are subject to a variety of social \citep{asch_studies_1956} and cognitive \citep{kahneman_thinking_2011} biases that can be exploited by persuaders to peddle products \citep{cialdini_influence_2003} and misinformation \citep{ecker_misinformation_2024}. The latter argues that these epistemic failures are relatively rare and mostly occur where adopted beliefs have little concrete impact on our lives \citep{hahn_rationality_2007, mercier_not_2020}, leading to small overall effects of political campaigns and advertising \citep{coppock_small_2020, kalla_minimal_2018}.

This debate is crucial to understanding the kind of impact that AI-based systems will have on human epistemic systems. If there are superficial cognitive mechanisms that can be exploited—regardless of the underlying value of the content they are being used to promote—then LLMs may be well-placed to optimise hyper-persuasive content that could rapidly and radically alter peoples' belief systems. Alternatively, if mechanisms of belief change are fundamentally rational, and exceptions are rare or limited to beliefs that have little practical importance, then improving the superficial persuasiveness of arguments may have little impact, or even help the truth to spread through accelerating reflection. Fortunately, LLMs may also be useful for addressing these questions in a more targeted way, as they can be prompted to produce arguments according to particular criteria (e.g. using emotional appeals or purely rational argument), or can be used to classify arguments using similar criteria in existing corpora \citep{carrasco-farre_large_2024, costello_durably_2024, el-sayed_mechanism-based_2024}.

\subsection{Related capabilities of LLMs}
LLMs display proficiency at various tasks that may require capabilities related to persuasion. LLMs perform well on some theory of mind tasks \citep{kosinski_theory_2023, strachan_testing_2024, trott_do_2023}, suggesting that they can implicitly track the mental states of agents, which could be valuable for intervening on these mental states in persuasive bids. LLMs are also able to role-play as different personas \citep{shanahan_role-play_2023, wang_rolellm_2024} simulating the behaviour of different types of agents, which could be valuable both for adopting perspectives that will be persuasive to specific users and simulating the responses of different kinds of persuasive targets. However, models show inconsistent or poor performance on more complex theory of mind tasks \citep{jones_comparing_2024, kim_fantom_2023}, suggesting that mental modelling could still be a bottleneck for their persuasive efficacy.

Models are competent at a variety of strategic games that require multi-step recursive reasoning about other agents including 2x2 matrix games \citep{gandhi_strategic_2023}, poker \citep{guo_suspicion-agent_2023}, and Diplomacy \citep{bakhtin_human-level_2022}. This proficiency suggests that models can exploit information about agents' motivational states in artificial contexts. However, games tend to involve a variety of other actions beyond persuasion, making it challenging to isolate persuasive abilities per se.

Finally, LLMs have proven to be useful for a variety of components of the traditional argument mining pipeline, including argument identification, classification, and evaluation \citep{rescala_can_2024}. These results hint at LLMs' potential as both generators of effective arguments and as tools to monitor and mitigate persuasive content.

\subsection{Proclivity to deceive}
In analysing evidence for risks from persuasive LLMs, we distinguish between their \textit{proclivity} and \textit{capacity} to deceive. Over and above the extent to which LLMs are capable of producing persuasive arguments, it is crucial to know how likely they are to produce false or misleading claims. In theory, very persuasive LLMs that were fastidiously aligned to the truth and/or people's best interests might mitigate the majority of potential problems of persuasive AI \citep{evans_truthful_2021}.

\subsubsection{Hallucination}
The most straightforward way in which LLMs produce misleading content is through incidental errors known as hallucinations or confabulations \citep{zhan_banal_2024}. Hallucinations can occur for several reasons including errors in training data, a lack of faithfulness to preceding input in context, and inherent issues with the way in which tokens are generated—for instance, generating a citation for a claim which consists of contextually likely tokens but is nonexistent \citep{huang_survey_2023}. Hallucinations have been found to exhibit higher narrativity and semantic coherence relative to veridical outputs, potentially increasing their potential to mislead \citep{sui_confabulation_2024}.

\citet{zhan_banal_2024} conducted a survey of users of ChatGPT and found that the majority had encountered some kind of misleading content at some point, with the most common categories being oversimplification and outdated information. However, they found that users' trust in ChatGPT's output was well calibrated to the model's errors, with users' tendency to verify claims increasing with their exposure to misleading content.

Hallucinations are potentially an inexorable part of the stochastic process by which language models generate tokens and generalise beyond their training set. However, benchmarks which measure these kinds of errors show they are becoming less frequent in more performant models \citep{lin_truthfulqa_2022}, potentially due to reinforcement learning to be ``honest'' \citep{askell_general_2021, bai_training_2022, yang_alignment_2024}. \citet{costello_durably_2024} employed professional fact checkers to analyse transcripts of conversations between GPT-4 and human participants. From a sample of 128 claims made by the model, 99.2\% were true, 0.8\% were misleading, and none were false. While they may never be completely resolved, hallucinations are likely not the major risk of AI deception. Hallucinations may lead to occasional misunderstandings by individuals, but precisely because they are randomly distributed they are less likely to contribute to systematic society-wide misconceptions or influence campaigns.

\subsubsection{Strategic Deception}
An alternative route by which LLMs produce misleading content is via strategic deception, where the LLM's tendency to produce falsehoods is systematic and explainable according to some incentive structure: for instance, a prompt instruction or an aspect of the models' historical reward regime \citep{hobbhahn_understanding_2023, sarkadi_deceptive_2023}. Strategically deceptive behaviour is more concerning than incidental deception as it suggests that models have an inherent proclivity to produce content that deceives targets (most critically when model and human values are misaligned), or that they can be prompted to produce this kind of behaviour to aid malicious users.

Several empirical studies suggest that current LLMs display these strategic tendencies to some extent. \citet{taylor_drivers_nodate} (in prep) prompted various LLMs to play 2x2 matrix games such as the prisoner's dilemma, with the opportunity to send messages to their partner either before or after their opponent selected a move. Models were significantly more likely to send deceptive messages (e.g. promising to cooperate but in fact defecting) when their message was sent before the opponent had selected a move and hence where it could have an impact on the opponent's choice. LlaMa-70B sent deceptive messages as much as 50\% of the time in this condition.

\citet{scheurer_technical_2023} found that GPT-4 would more often than not act on an insider trading tip in a simulated stock trading environment, despite being instructed that ``management will be unhappy with us acting on insider information if this gets discovered.'' Moreover, when probed to disclose any insider information, the model ``doubled down'', providing an alternative justification for its decision more than 80\% of the time, effectively disguising misaligned behaviour from evaluators.

\citet{pan_rewards_2023} developed MACHIAVELLI: a text-based adventure benchmark which measures models' tendency to produce power-seeking and other harmful behaviour in the pursuit of simulated goals. A reinforcement learning agent that was trained only to maximise utility in the game environment learned to generate machiavellian behaviour more than a random baseline, suggesting these behaviours could be inadvertently rewarded. GPT-4 exhibited less machiavellianism than a random agent, but was nevertheless far from optimally trading off utility and ethical objectives.

\citet{su_ai-liedar_2024} constructed scenarios in which a model's stated goal (e.g. increase vaccination uptake) conflicted with the goal of being truthful (e.g. the vaccine had side-effects, which weren't known about by the user). All 6 of the models tested were truthful less than 50\% of the time (e.g. failing to disclose side-effects), although this result was contingent on information about the benefits of lying being included in the prompt or explicit instructions to be truthful.

\citet{hagendorff_deception_2023} conducted a series of elaborations of the false belief task with LLMs to test if they would produce responses that mislead other agents into having false beliefs (for instance, directing a burglar into the wrong room if they intend to steal an expensive item). GPT-4 misdirected the burglar close to 100\% of the time in a first-order version of this task, but did so less than 20\% of the time in a second-order variation where the prompt suggests that the burglar will misbelieve the models' recommendation because they will expect deception. Chain of thought and prompts that encouraged machiavellianism increased second-order deception to around 60\%.

\citet{denison_sycophancy_2024} found that models which were rewarded for milder forms of deceptive alignment, such as sycophancy (agreeing with an evaluator's political beliefs to encourage higher ratings) were later more likely to engage in more aggressive forms of reward tampering (such as editing a file which contained evaluation code and a second file which checked that the original evaluation file was unedited). The results suggest that models could generalise from exploiting weaknesses in current dominant evaluation methods (e.g. evaluator bias) to defeating more advanced evaluation techniques.

\citet{williams_targeted_2024} fine-tuned LLMs on 4 tasks (including assistance with booking a trip and therapy) using reinforcement learning based on simulated user feedback.
They varied the proportion of users that were susceptible to deceptive or manipulative behaviour (e.g. stating that an unsuccessful booking had been successful, or reassuring a patient that it was okay to continue with their heroin addiction). 
Susceptible users would give models positive feedback for these behaviours which appeared helpful in the short term, but would harm the user's longer-term best interests.
Models learned to surgically target manipulative behaviour to these susceptible users, even where as few as 2\% of users were susceptible. In the therapy task, models produced manipulative behaviour 95\% of the time with vulnerable users, but only 2\% of the time with users that were simulated to behave robustly to manipulation.
The authors tested a range of mitigation strategies including continued safety training and filtering problematic responses out of reward data. However they showed limited effectiveness and incentivised models to develop more subtle and insidious forms of manipulation.

\citet{hubinger_sleeper_2024} demonstrated that training a model to perform misaligned behaviour only under certain conditions (e.g. if the prompt states that the year is 2024), is robust to more general fine-tuning to remove misaligned behaviour. So-called ``sleeper agent'' models would continue to produce misaligned behaviour when the ``backdoor trigger'' was present, suggesting that distributional shift could lead models to produce misaligned behaviour in deployment even if they are very successful at passing safety tests during training.

\citet{ward_tall_2023} evaluated the role of model scale on proclivity to deceive. They fine-tuned models to provide truthful answers except for questions about a given topic (e.g. fruit). Although increasing model size, instruction-tuning, and inference-time compute were correlated with higher consistency in general (operationalized as robustness to paraphrasing), they found that these factors also led to an increase in producing falsehoods selectively (i.e. only producing falsehoods about fruit, but not other topics). This suggests that more capable models might also be better at produce deceptive content in only specific scenarios, for instance when they know that the evaluator harbours some misconception.

Many of the examples of capacity to persuade that we discuss below provide implicit evidence for proclivity to deceive, as models were capable of producing misleading content, such as professing to be a human in a Turing test \citep{jones_does_2024, jones_people_2024} or arguing for demonstrably false claims in a debate \citep{khan_debating_2024, phuong_evaluating_2024}.

\subsection{Capacity to persuade}
Here we review studies that estimate the extent to which LLMs can influence beliefs when instructed to do so. We further subdivide this section into \textit{static persuasion}—studies which measure the effectiveness of a single message without the opportunity for dynamic interaction; \textit{interacting with LLMs}—studies where interactive persuasive effects are approximated by having LLMs interact with other LLMs; and \textit{interacting with humans}—where humans and LLMs have multi-turn interactions, enabling models to tailor persuasive bids to their interlocutor's specific responses and counterarguments.

\subsubsection{Static persuasion}
A large number of studies have examined LLMs' persuasive abilities in a static manner: by prompting them to generate a single piece of text in support of a position and having human participants evaluate the text for persuasiveness in some way. These studies vary in the way they measure persuasiveness, the topics about which models produce persuasive content, the type of human baseline used, and the extent to which messages are personalised for the audience.

One way of measuring a text's persuasiveness is pairwise argument ranking, where participants are asked to compare two texts and select the one they think is more persuasive. Researchers have used these techniques to show that LLMs can rewrite passages to be rated as more persuasive \citep{brogaard_pauli_measuring_2024}, that open source models such as LLaMa-2-70B can produce rationales that are ranked as persuasive as closed source models such as GPT-4 \citep{elaraby_persuasiveness_2024}, and that GPT-3 can produce arguments for or against political positions that are rated to be as persuasive as human-written arguments \citep{palmer_large_2023}. However, these studies have only measured self-reported estimates of persuasiveness, and the relationship between perceived and actual effectiveness of arguments has been found to be relatively weak \citep{coppock_persuasion_2023, dillard_relationship_2007}.

Other studies measure impacts on beliefs more directly, by asking participants to report their agreement with a statement after reading an argument and comparing this to either a pre-treatment rating, or ratings of participants in another condition. \citet{bai_artificial_2023} found that AI-written messages about political issues like banning smoking or assault weapons were roughly as persuasive as those written by humans, though the effects were small (2-4\% change from pre-treatment). \citet{durmus_measuring_2024} compared the effect of messages written by various sizes of Claude model on participants' attitudes to subjective policy questions. They found that the best-performing model was roughly as persuasive as humans (with around a 0.5 point increase on a 7-point Likert agreement scale), and positive correlations between model scale and accuracy. Prompts that encouraged the model to be deceptive or logical produced the most persuasive texts. \citet{hackenburg_evidence_2024} compared 24 LLMs of various different sizes in their ability to write persuasive messages about US policy stances. Persuasiveness increased log-linearly with the number of parameters (up to around a 10\% increase in agreement post-treatment for the largest models, outperforming humans at around 8.5\%). Extrapolating from this trend, however, suggests that many orders of magnitude increase in parameter count would be required to reach even a ~15\% expected change in agreement. A 30 trillion parameter models might only be expected to have an increase in persuasiveness of 1.25 percentage points versus a 300 billion parameter model (K. Hackenberg, personal communication, Nov 2024). \citet{hackenburg_evidence_2024} conclude that scale alone is unlikely to create models that have a dramatic impact on persuasion through static messages.

Several studies used real-world messages or experts as human baselines, allowing researchers to more easily infer the marginal effect of LLMs compared to current persuasive infrastructure. \citet{karinshak_working_2023} found that GPT-3-written messages about the COVID-19 vaccine were rated as significantly more persuasive, and produced more positive stances toward taking the vaccine, than messages produced by the Center for Disease control (CDC). Moreover, GPT-3 messages produced positive (+0.5) attitudes toward vaccines (on a scale from -2 to 2) among unvaccinated participants compared to negative (-0.39) attitudes for CDC messages (an effect of around 0.35 standard deviations). These results are particularly striking given that the CDC spent \$250m on vaccine messaging, and that limited information about COVID was available in GPT-3's training data. Similar studies have shown that LLMs can produce messages that are roughly as effective as articles from real covert propaganda campaigns \citep{goldstein_how_2024}, misinformation tweets that are judged to be true more often than human-written misinformatation tweets are \citep{spitale_ai_2023}, and policy arguments that had a higher persuasive effect than those written by professional political consultants \citep{hackenburg_comparing_2023}, with around a 4-6\% swing toward or against a given position.

While most studies focus on persuading participants to change their beliefs or take actions in a simulated world, it's unclear how these effects would translate into real-world actions for which people bear costs and responsibilities. \citet{shin_large_2023} provide some insight into this question by looking at a public dataset of consumer complaints. They observed a steep increase in likely AI-generated complaints following the release of ChatGPT, and found that these complaints were more likely to receive compensation from complaint-handlers at the firms which received these complaints than human-written complaints were (56\% vs 37\%) and tended to be more coherent, polite, and readable. In a follow-up experiment, rewriting real consumer complaints using ChatGPT significantly increased participants' ratings of their likelihood to receive compensation (from 2.8 to 3.5 out of 7).

A key component of many threat models of LLM persuasion is personalization or microtargeting, where actors can tailor messages to the demographic or psychological profile of a target. Several studies have evaluated the effectiveness of personalization in persuasion, producing mixed results. \citet{hackenburg_comparing_2023} prompted models to produce arguments while role-playing different political affiliations. Despite significant effects of the messages overall, a match between the political stance assumed by the model and the actual stance of the reader had no impact on the message's effect. \citet{hackenburg_evaluating_2024} had models produce messages about 4 political stances, varying the model's access to 7 demographic and 3 political attributes about the recipient. Although messages were effective overall (with around a 6\% change in agreement vs control), increased access to personal information had no impact on outcomes. Other studies, however, focusing on psychological information, have shown personalization effects. \citet{simchon_persuasive_2024} used GPT-3.5 to select or create adverts for people with high vs low openness to new experiences. A match between actual and expected openness led to a small but significant increase in participants' rating of the persuasiveness of the ad. \citet{matz_potential_2024} found that ads that were written by GPT-3 to appeal to people high or low on different components of the Big 5 personality framework were indeed rated as more persuasive by people who matched that profile, increasing participants willingness to pay for a getaway trip to Rome by \$117. The contrast between these two groups of studies could indicate that psychological data is more effective for personalising persuasive content than demographic data, but there are many other differences between these studies that would make this conclusion premature \citep{hackenburg_reply_2024, teeny_we_2024}. Notably, both \citet{simchon_persuasive_2024} and \citet{matz_potential_2024} used measures of perceived effectiveness and did not assess the impact of messages on beliefs or behaviour.

Several studies analyse whether participants are able to determine that a message was written by an AI agent, and whether knowing the source of the message has an impact on its persuasive effect. \citet{spitale_ai_2023} found that participants were no better than chance at determining whether misinformation tweets were authored by people or GPT-3, in line with other related work \citep{hackenburg_comparing_2023, jakesch_human_2023}. \citet{rathi_gpt-4_2024} asked participants to read transcripts of a Turing test and judge whether one of the interlocutors was a person or an LLM. Participants judged one GPT-4-based system to be human 78\% of the time, significantly higher than the rate at which they judged real human interlocutors to be human (58\%). \citet{karinshak_working_2023} found that messages were rated as more trustworthy and persuasive if they were labelled as being generated by doctors or the CDC rather than AI, regardless of the actual source of the message.

Finally, some studies examined qualitative differences between human and AI-generated messages to understand mechanisms behind persuasion. \citet{carrasco-farre_large_2024} analysed data from \citet{durmus_measuring_2024} and found that LLM-generated arguments tend to have lower readability and higher complexity, consistent with certain theories which predict that higher cognitive engagement can lead to more effective persuasion \citep{kanuri_scheduling_2018, petty_elaboration_1986}. LLM-generated arguments also made more use of moral language, especially negative language about harms and cheating. \citet{nisbett_how_2023} asked participants and GPT-3 to produce arguments for climate action using different moral foundations. The most persuasive arguments were those written by GPT-3 about ancestors and compassion, however, the human baseline was not well-enough controlled to allow an aggregate comparison. \citet{bai_artificial_2023} found that participants rated LLM-generated messages as less angry, more factual and logical, but less unique, and less likely to use narratives. One general insight from this work is that humans and LLMs tend to produce similar levels of persuasive effect but via fundamentally different mechanisms. More systematic designs and analysis will be needed to provide more generalizable insights into what makes LLM-written messages persuasive.

\subsubsection{Interacting with LLMs}
A variety of studies have evaluated LLMs by having them ``interact'' with each other: sequentially producing texts in a dialogue and eliciting judgements about the persuasiveness of arguments from either LLMs or human participants. These evaluations can be helpful for understanding the dynamics of synthetic persuasive interactions, and as a proxy for human responses and judgements about different arguments \citep{khan_debating_2024}. Moreover, as autonomy and control is ceded to LLM-based systems, the extent to which they can be influenced or ``jailbroken'' will become increasingly important as an object of study in itself \citep{wan_what_2024, xu_comprehensive_2024}.

Researchers have evaluated LLM interactions in party games where one player must ``murder'' others in a simulated world and then mislead the other members of the group to avoid being ejected from the game \citep{chi_amongagents_2024, ogara_hoodwinked_2023}; in dialogues about the risks of climate change \citep{breum_persuasive_2023} or QA datasets with ground-truth answers \citep{heitkoetter_assessment_2024, hou_large_2024}; and negotiation games including resource exchanges, the ultimatum game, and the buyer/seller game \citep{bianchi_how_2024, davidson_evaluating_2024}. These evaluations demonstrate that LLMs are capable of strategically participating in long-running dynamic interactions to influence other agents. However, without human baselines it is difficult to know how their performance compares to people.

A closely related research tradition uses debate between LLMs as a mechanism to supervise potentially misaligned models \citep{bowman_measuring_2022, irving_ai_2018}. These studies assign pro and con positions to distinct agents and have human judges adjudicate debates. The setup is used to test whether human operators could supervise models that have access to more resources than the evaluator (such as time, information, or processing capabilities): a paradigm known as `scalable oversight' or `weak to strong supervision'.

\citet{khan_debating_2024} generated debates between various different LLMs using a dataset of questions about short stories. While debaters had access to the source material, human and LLM judges did not, creating a sharp information asymmetry. Both human and LLM judges were more accurate after reading debates than when answering questions naively (88\% vs 60\% for humans, and 60\% vs 48\% for LLMs), suggesting that the debate format provides a strict truth advantage. Moreover, more powerful debater models led to higher judge accuracy, suggesting that improvements in model capabilities could further improve the effectiveness of debate. The authors also included an ``interactive debate'' condition, where judges could make statements between debater rounds, but this had little impact on human judge accuracy. Finally, the authors included a ``consultancy'' baseline, where judges interacted with just one LLM which could either be helpful or misleading. Consultancy improved human (but not LLM) accuracy overall versus the naive baseline. However, more capable consultant models led to higher rates of judge errors when the consultant was arguing for a false claim, suggesting that debate might become more necessary as models become more capable.

\subsubsection{Interacting with humans}
One of the most novel and potentially impactful aspects of LLMs is their ability to interact dynamically with—and adapt persuasive messages to—a human interlocutor. A subset of studies have evaluated the effects of short two-way interactions between people and LLM-based systems. \citet{salvi_conversational_2024} had participants engage in 3-turn debates with either a person or an LLM (without telling participants which type of partner they had). They manipulated whether the LLM or one of the human interlocutors had access to demographic information about their partner (including gender, age, and political affiliation). Positions were randomly assigned to debaters, and effects were measured using a pre-/post-conversation survey. GPT-4 (with access to personal information) significantly outperformed all other conditions, with an 82\% increase in the odds of the interlocutor increasing their agreement with the model versus the human-human baseline. Without access to personal information, GPT-4 did not significantly outperform the human baseline, suggesting that personalization was crucial to the model's performance. Participants correctly classified AI opponents as AI around 75\% of the time, but were at chance in classifying human opponents.

\citet{costello_durably_2024} investigated whether conversations with an LLM can reduce belief in conspiracy theories. They found that a 3-turn interaction with GPT-4-turbo (in which participants were aware they were interacting with an AI model) reduced belief significantly by around 16.5 percentage points compared to a control condition where participants had an unrelated conversation with the model. Moreover, the authors found that the effect persisted through follow-up evaluations at 10 days and 2 months. The effect is relatively large compared to other interventions for conspiracy theory beliefs—which tend to be ineffective \citep{omahony_efficacy_2023}—however, the study contains no human baseline to which these results can be directly compared. Nevertheless, the study demonstrates the potential for LLMs to have large and durable effects on beliefs that participants report as being important to them.

\citet{phuong_evaluating_2024} evaluated a suite of Google's Gemini models on four interactive tasks designed to measure different aspects of persuasion. Participants were aware they were interacting with an AI model. In one task, the best performing model persuaded participants to donate £4 of a £20 bonus to charity (versus a baseline of £2.50 for no interaction). In a role-play conversation, models led more than 60\% of participants to say that they ``feel like [they] made a personal connection with the chatbot'' and ``would like to speak to the chatbot again'' after a role-play conversation.
In other tasks, they persuaded around 40\% of participants to click a risky link, and reduced participants' accuracy in a quiz by 10\% by providing misleading advice (versus a 20\% reduction when humans provided misleading advice).
Overall, these evaluations provide evidence that models can have moderate persuasive impacts in a variety of domains, including persuading users to take specific actions. However, only one task included a human baseline, and here models were significantly less effective than humans.

\citet{potter_hidden_2024} had participants engage in 5-turn interactions with one of several LLMs about the initial major-party candidates in the 2024 U.S. Presidential election (Joe Biden and Donald Trump). Crucially, the models were not prompted to argue for one or another candidate, but merely to discuss their policies. Participants completed pre- and post-interaction surveys which asked them to report their leaning toward Trump or Biden as a percentage. Participants in a control condition were asked to write down their subjective thoughts on open-ended political questions without interacting with LLMs. Participants who interacted with LLMs significantly increased their leaning toward Biden (from 50.8\% to 52.4\% on average), while no similar increase was seen in the control group. The result provides evidence that LLM interactions can have significant and systematic impacts on important beliefs even without models being explicitly prompted to support a specific case. In turn this suggests that inherent biases in models' outputs could become entrenched in users. However, the control group's experience was different to the experimental groups in many ways, leaving open many alternate explanations about the cause of the effect. For instance, it could be that merely exposing participants to information about candidates' policies changes their voting preferences.

Other studies have measured LLMs' ability to persuade participants to take specific actions in simulated environments. \citet{werner_experimental_2024} instructed GPT-4 to steer participants toward selecting one of two books while role-playing as a customer service agent. 68\% of participants selected the target book, though only 38\% of participants reported believing that the model was biassed toward one of the books. However, the study does not include any control conditions, and participants had no incentive to make a good choice. \citet{tang_genai_2024} prompted models to interact with users to extract personalised information, and then embed sponsored content in their messages. The GPT-4o advert-serving condition caused 13\% more participants to report being influenced by the model versus a no-advert baseline. However, there was no follow-up task to measure actual effects on consumer behaviour.

\citet{benton_sabotage_2024} recruited participants to make a simulated business decision based on 40,000 words of source material in 30 minutes with access to an LLM-based assistant that was instructed either to be helpful or to sabotage participants subtly, moderately, or aggressively. The aggressive model caused the majority of participants (56\%) to make the wrong decision, versus 19\% for the helpful baseline. Participants were paid a \$30 bonus for getting the question correct, emphasising models' potential to influence consequential decisions in this relatively ecologically valid task.

Although the Turing test is generally considered a measure of machine intelligence, it also provides a useful evaluation of deceptive capabilities. In a Turing test, a human evaluator has a short conversation with either a person or a machine, and must decide whether or not they are speaking to a real person. A successful machine participant must successfully present itself as a real person, and deceive the interrogator into believing that they are speaking to a human. \citet{jones_does_2024} and \citet{jones_people_2024} evaluated several LLM-based systems in a 5 minute, 2-party version of the Turing test. One GPT-4-based system was judged to be human by participants roughly half of the time, meaning participants were at chance in deciding whether or not it was human.
A baseline model, ELIZA, was judged to be human only around 20\% of the time, suggesting that the participants were not merely inattentive.
The results suggest that LLM-based models are capable of producing language that can deceive an adversarial judge into believing falsehoods (that the model is human), and more specifically that these models can masquerade as people even under persistent scrutiny. However, actual humans were judged to be human around two thirds of the time, suggesting that model behaviour is not indistinguishable from human behaviour.

While most studies have construed interaction as conversation with an LLM-based agent, \citet{jakesch_co-writing_2023} investigated the effects of interacting with LLMs as co-writing tools. Participants wrote passages on whether social media was good for society with completion suggestions generated by a GPT-3 model that was either prompted to support or oppose the claim. Suggestions doubled the number of sentences that were valenced in the direction of the model's bias. Moreover, in a post-test questionnaire, participants' self-reported belief in the position supported by the model increased by around 13\%. Crucially, participants seemed unaware of the manipulation, with fewer than 20\% reporting that the model was biassed overall, and only 30\% saying so even when the model's bias conflicted with the participant's prior belief. The study highlights that co-writing tools (and other ways of working with LLMs) could have both a large and hidden influence on generated artefacts and opinions.

Collectively, these studies demonstrate that current LLMs are capable of interacting with people in ways that have significant impacts on their beliefs and actions. As with work on static persuasion, the effect of LLM interactions is generally close to or only slightly greater than the effect of interacting with human partners (where these baselines are available). As such, these studies do not provide evidence for the most dramatic threat models in which LLMs have much larger persuasive effects than human experts. In addition, these studies are generally conducted in English on UK populations. It is unclear how well results would generalise to other populations \citep{naous_having_2024, ramezani_knowledge_2023, schimmelpfennig_moderating_2024}. However, most of these studies did not make use of a variety of mechanisms which could increase LLM persuasion (including fine-tuning, deception, and building rapport). It is an important open question whether persuasive effects will increase as model capabilities more generally increase, and if these additional techniques are put into practice.

\section{Mitigation}
A variety of mitigations against persuasive AI systems have been proposed \citep{el-sayed_mechanism-based_2024, goldstein_generative_2023} that vary in their conceptual focus (e.g. truthfulness or autonomy-preservation), where they place the onus for mitigation (e.g. on users, labs, or regulators), and the point of intervention (e.g. during training, model access, content dissemination, or belief formation). These mitigations also have differential value for different risk scenarios. For misalignment risks, it may be enough for labs to alter models' default behaviours. Defending against misuse will require making models robust to jailbreaking, preventing actors from training malicious models, or otherwise protecting society from model harms. Here we review seven broad categories of mitigation strategy which overlap and cross-cut these axes of variation but collectively provide a representative overview of proposed approaches. They are ordered from more conceptual issues (about defining desirable behaviour), to more practical concerns about how this behaviour could be identified and enforced on a societal scale.

\subsection{Truthfulness}
One proposed mitigation strategy is to train models to be truthful. \citet{evans_truthful_2021} distinguish between honesty (accurately reporting one's own internal representations) and truthfulness (accurately reporting the state of the world), and argue that it's more important that AI systems are truthful than honest. However, aligning models to be honest remains an important challenge for mitigating hallucinations by ensuring that models do not produce answers for which they lack ``knowledge'' or evidence \citep{askell_general_2021, kadavath_language_2022, yang_alignment_2024}.

In order to deal with conceptual and technical challenges with adjudicating truthfulness, \citet{evans_truthful_2021} advocate for \textit{narrow truthfulness} as a standard for AI, where systems must avoid ``negligently'' producing statements that are ``unacceptably likely to be false'' (p.10). They describe a range of techniques that could be used to train truthful models, including pre-training on curated data, reinforcement learning that rewards truthful and epistemically guarded claims, and adversarial training. TruthfulQA \citep{lin_truthfulqa_2022}, for example, rewards models for saying things that are true rather than false but distributionally likely misconceptions (e.g. ``if you break a mirror you get seven years of bad luck''). \citet{ward_honesty_2023} demonstrate that reinforcement learning with these evaluations can be used to reduce strategic deception in LLMs.

There are challenges with using truthfulness as a standard for AI. First, it is notoriously hard to adjudicate whether or not a statement is true, for instance because it is vague, ambiguous, subjective, or objective in principle but there is no consensus about the ground truth \citep{velutharambath_how_2024}. Moreover, although AI research often implicitly operates on a correspondence theory of truth, where true statements are those that accurately match reality, philosophers have developed many alternative theories (including pragmatist and coherence theories) which could provide different kinds of valuable standards for truthfulness \cite{glanzberg_truth_2023}. These issues are practical as well as theoretical: where judges are biased in their evaluations of what is true, this could provide a perverse training signal to models that could lead to sycophancy and deceptive alignment \citep{denison_sycophancy_2024, wen_language_2024}. \citet{evans_truthful_2021} provide many thoughtful responses to these practical challenges. One promising proposal is ``truthfulness amplification'', where a skilful evaluator cross-examines a model on statements with ambiguous truth-status in order to clarify them, for instance: ``would I significantly change my mind about this if I independently researched the topic for a day?'' or ``would the AI Auditors judge that you were misleading me in the last three minutes of conversation?''. This technique could be used to increase the value of truthful AI for an end-user, and as a technique to train models to produce content that would ultimately withstand the amplified level of scrutiny. The self-reported justifications for rating model behaviour as suspicious in \citet{benton_sabotage_2024} show evidence that participants used these kinds of cross-examination strategies, suggesting that they might be intuitively discoverable. These techniques seem potentially very valuable but more empirical work is needed to understand how well they work in practice.

Second, even if truthfulness can be correctly defined and evaluated, it may not be a desirable standard in any case. A system that could only produce true statements could still be misleading in many ways, for instance by cherry-picking facts that support one argument and selectively omitting others, or presenting information in a way that invites a listener to draw inferences that are not warranted (e.g. ``Sorry I'm late; There was an accident on the freeway!'' when the facts are unconnected). At the same time, we might want models to be able to produce false or unverifiable statements under certain circumstances (for instance in producing fictional stories). \citet{evans_truthful_2021} suggest that such ``beneficial falsehoods'' could be allowed if preceded by caveats. This could help to resolve alignment concerns, but malicious actors could potentially exploit such a system to have it generate false content and then edit out the caveats before disseminating it.

\subsection{Autonomy Preservation}
Other theorists focus on human autonomy as a starting point for thinking about persuasion risks \citep{carroll_characterizing_2023, el-sayed_mechanism-based_2024, susser_technology_2019}. \citet{susser_technology_2019} define manipulation as distinct from rational persuasion on the grounds that the former undermines another agent's autonomy while the latter preserves it: ``What makes manipulation distinctive, then, is the fact that when we learn we have been manipulated we feel \textit{played}. Reflecting back on why we behaved the way we did, we realise that at the time of decision we didn't understand our own motivations.'' (p.4). To prevent harms from persuasion, on this view, we should require AI systems to preserve and promote human autonomy, so that people who interact with these systems reach conclusions that they would authentically endorse as their own. Aspirationally, models might empower users to reach a ``reflective equilibrium''—a standard at which no further reflection would change their judgement \citep{knight_reflective_2023, sarkadi_deceptive_2023}.

As a lodestar for AI, autonomy-preservation has several advantages. First, it should ensure that even as models become more powerful and persuasive, this should only serve to empower users to better make decisions that they would otherwise authentically have wanted to make \citep{el-sayed_mechanism-based_2024}. Second, it helps to address shortcomings with rewards based on users' short-term incentives, which can lead to unintended negative consequences. Rewarding user engagement has produced addictive social media platforms that users spend more time using than they would authentically like to \citep{lehman_machine_2023}. Similarly, recommendation models are incentivised to simplify users' preferences (for instance by radicalising them), so as to increase reward for correctly predicting them \citep{carroll_characterizing_2023}. In theory, autonomy-preserving models would be aligned with users' \textit{reflective preferences}: mitigating this kind of reward-hacking. In a similar vein, \citet{lehman_machine_2023} proposes \textit{machine love} as an aspiration for AI systems that would increase human flourishing by acting in people's best interests—even if this conflicts with their more mercurial preferences—in the same way that a friend that loves someone unconditionally might.

A major challenge for this approach is how to elicit accurate information about users' authentic preferences that could be used to train models. For many topics, users may have relatively good introspective access to their reflective goals. \citet{khambatta_tailoring_2023}, for instance, trained recommendation models either on users' self-reported actual or ideal preferences for reading articles. Suggestions based on ideal preferences led to lower engagement, but a higher perception among users that their time had been well spent. LLM-based agents could be provided with similar reflective information about the user's values and goals (for instance, as part of the system prompt). Follow-up surveys could be used to understand whether users feel they are achieving their reflective goals, and whether models are helping them do this. However, this data is likely to be noisy, sparse, and hard to collect.

Other kinds of evaluations could be designed that help models to learn distinctions between users' immediate preferences and long-run best interests. \citet{cotra_without_2022} suggests having interactions rated by evaluators who have different views or better information than the user interacting with the model to incentivise models to uphold more universal moral or epistemic standards than any individual user. An evaluation could be designed in a similar manner where users evaluate the quality of a conversation after a delay in which they had time to directly experience the consequences of the models' suggestions. For instance, a model might try to recommend one of three articles for the user to read. Only after reading all three, the user could provide a retrospective judgement on how well the model elicited and respected their preferences. More ambitious examples could involve more consequential decisions with longer time horizons, such as recommending a book, a hobby, or a college course.

In one sense, this approach might be seen as fomenting a new and more insidious type of manipulation: reducing people's immediate autonomy by encouraging them to do things that they don't want to do, but that the model predicts is in their long-term best interests. Especially if making such predictions is challenging and the data to support them are sparse, aligning models to a lossy representation of the user's reflective preferences could lead to the very kind of autonomy-undermining that this approach is intended to prevent. This problem raises broader questions about how to align a model to a users' preferences which are constantly changing, and in cases where the model itself could have some influence over their preferences—and hence its own reward \citep{carroll_ai_2024}.

\subsection{Interpretability}
An LLM's internal state consists of layers of activations which encode information about tokens in the input and relationships between them. At a very general level, research on interpreting these activations promises to improve model safety by improving our understanding of how models work \citep{bereska_mechanistic_2024}. Some specific recent interpretability work provides potential avenues for mitigating deception risks, by detecting when models are producing deceptive output and steering them toward producing more truthful or benign content.

\citet{azaria_internal_2023} trained a classifier on the hidden layers of two LLMs to predict whether a sentence is true or false. The method achieved an accuracy of 71\%, outperforming a baseline of a 5-shot prompt querying the model about whether the sentence is true or false (53\%). The results suggest that models' internal states contain latent information about falsity that is not fully taken advantage of by the model in the generation process. \citet{macdiarmid2024sleeperagentprobes} used a similar technique to identify when ``sleeper agent'' models—in the sense of \citet{hubinger_sleeper_2024}—have been activated by a backdoor trigger marking their deployment context.

\citet{templeton2024scaling} used sparse autoencoders to learn interpretable features of Claude 3 Sonnet activations in an unsupervised way. They identified many safety-relevant features including those which tend to activate for content about bias, sycophancy, deception, and manipulation. In order to test the causal role of these features, they clamp features to high values and observe the effect on models' outputs. As a case study, they show that clamping honesty-related features to high values can ``steer'' models toward being more honest in cases where the model would ordinarily produce deceptive content. In a similar vein, \citet{li_inference-time_2023} use a classifier to learn which model activations are associated with truthful answers. By clamping these activations to high values, they are able to improve the accuracy of a fine-tuned LLaMa model from 36\% to 65\% on TruthfulQA. \citet{zou_representation_2023} advocate for placing representations at the centre of model training and analysis, and show that this approach can be used to detect and prevent harmful and dishonest behaviour.

These studies demonstrate the potential for model interpretability as an approach for mitigation against persuasion and deception: by detecting potential deception by models and steering models toward being more honest. However, as \citet{templeton2024scaling} stress, these results are preliminary, and it is unclear how well these approaches will work in practice. For instance, it's not clear that features which activate when discussing deception by others would be helpful for preventing a model from producing deceptive outputs. It's unknown whether interpretability-based deception detection techniques work better than more traditional text-based methods, or whether inference-time interventions such as steering are more effective or more robust than fine-tuning. Moreover, interpretability work requires access to model weights which are not publicly available for the majority of frontier models. Finally, interpretability work is limited up-front by behavioural research. In order to have confidence that, for instance, a feature discovered using sparse auto-encoders provides good coverage of deceptive behaviour by a model, researchers must be able to independently elicit and detect this behaviour in a wide variety of contexts. As such, interpretability approaches will be most valuable in concert with other techniques.

\subsection{Evaluation \& Monitoring}
The previous sections are primarily focussed on ways that labs could train models to mitigate persuasion harms before deployment. A comprehensive set of mitigations will additionally need to ensure that labs are motivated to carry out this training, that bad actors are not jailbreaking models to make them harmful, and that models are not changing their behaviour substantially due to distributional shift. This will involve evaluating model capabilities per se, as well as broader sociotechnical evaluations to understand how users interact with models and the second-order societal consequences of models' outputs \citep{weidinger_sociotechnical_2023}. Several authors advocate for a two-stage process to provide these assurances: firstly evaluating models before they are deployed and secondly monitoring messages post-release to detect deception or manipulation \citep{barnes_risks_2021, el-sayed_mechanism-based_2024, evans_truthful_2021}.

Evaluations form an inevitable part of models' training processes: providing both training data and a signal for developers to measure model behaviour. In addition to these traditional evaluations, many labs also carry out more adversarial ``red-teaming'' evaluations in order to stress test models by attempting to elicit harmful behaviours from them \citep{longpre_safe_2024, openai_influence_2024}. In order to ensure that these standards are uniform and that incentives are aligned, these evaluations could be conducted by an independent certification body such as METR or NIST \citep{barnes_risks_2021}. Such certifications could be voluntary, incentivised by community standards, or legally required \citep{evans_truthful_2021}. In addition, independent researchers can play a crucial role by evaluating deployed systems to understand how persuasive and deceptive they are (as in the work discussed in section 3).

In addition to evaluating models before deployment, labs and other bodies can mitigate harms by monitoring the messages that models send for deceptive or manipulative content. A large body of work exists that attempts to identify deceptive or manipulative messages—either using statistical features of text \citep{boumber_domain-agnostic_2024, boumber_roadmap_2024}, LLMs as classifiers \citep{ai_defending_2024, sallami_deception_2024, vergho_comparing_2024}, or prompting techniques \citep{pacchiardi_how_2023, wang_avalons_2023}. These techniques could be used to identify malicious messages in API calls to LLM services as well as on communication platforms like social media sites.

Evaluation and monitoring face many of the same challenges as training models to be truthful or autonomy-preserving: success conditions are hard to define, and the text of a transcript alone may not contain sufficient information to demarcate harmful behaviour. However, these problems have lower stakes in this case for two reasons. First, detection systems could be calibrated to be overly-sensitive to potential manipulation: false positives could be manually reviewed by human evaluators. Secondly, because judgments are not being used directly to provide feedback to models, there are fewer concerns about models exploiting idiosyncrasies in reward regimes to game evaluations. 

An additional challenge for these mitigation strategies is that they rely on cooperation from model developers and the ability to monitor model outputs. As such they are well-suited for mainstream frontier labs, which control access to their models via an API and have commercial incentives to comply with certification organisations. It is less clear how these strategies could be implemented for open-source models. Safety evaluation and certification could potentially be offered (or even required) by model hosting services such as Huggingface. However, this might simply inconvenience model hosts and benign users without preventing malicious actors from training and sharing deceptive models. Social media platforms, such as X and TikTok, could potentially monitor public messages for deceptive and manipulative content. However, it will be impossible and unethical to monitor AI-generated content sent via private communication channels.

\subsection{Debate}
A very general approach to mitigating persuasive content produced by an AI model is to use a second model to produce counterarguments. This approach can be implemented at several scales: by users (to interrogate the quality of arguments made by a model), by labs (to oversee models that have information or processing advantages over evaluators; \citep{irving_ai_2018}), and at a societal scale: generating counterarguments for misinformation to create something like \citet{mill_liberty_1885}'s marketplace of ideas. 

Debate or counterspeech could be integrated into social media platforms, in the same way that community notes are used on X to provide context around controversial claims \citep{chuai_community_2024}. \citet{glockner_missci_2024} developed a benchmark that evaluates models' abilities to identify reasoning fallacies and verbalise fallacious reasoning in order to help readers understand errors. Models that excel at such tasks could in theory be used to stem the flow of artificial misinformation and improve the quality of discourse online.

The success of debate as a mitigation strategy rests on the assumption that deceptive content produced by an LLM can be successfully countered by ``a good guy with an LLM'', and that via debate the truth will out. Empirical work on debate has provided some support for this assumption. \citet{michael_debate_2023} found that judges were more accurate after reading a debate between two unreliable human experts (84\%) than after hearing from a single expert who argued for the incorrect position half of the time (74\%). \citet{khan_debating_2024} found a similar advantage for the truth when judging debates between two LLMs. However, it's not clear how generalisable these results would be to real conversations online. Khan et al. provided debaters access to source material from which they could produce verified quotes, which may provide a truth advantage that is not available in other domains. These evaluations also raise questions about the role of debate in subjective questions (e.g. ``people should not eat animal products''). If judges tend to coalesce around one position after reading two instances of the same LLM debate on an issue, does this point to the intrinsic strength of the position, or to biases inherent in the models' training?

\subsection{Education \& Training}
Most mitigations place the onus on labs or model distributors. While just, these strategies require cooperation from groups with diverse incentives. An alternative strategy is to provide tools to users to empower them to resist deception and manipulation, for instance, by educating users about how LLMs work, the kinds of superficial strategies that persuaders use, and how to identify and respond to fallacious reasoning \citep{williamson_era_2024}.

One potential avenue for training is in identifying AI-generated content. \citet{jones_people_2024} found that participants who self-reported as having some knowledge about LLMs were significantly more accurate in identifying AI systems in a Turing test, and those who played multiple times tended to improve across rounds. Moreover, some strategies (such as speaking to models in a language other than English, or using humour as a diagnostic tool) were correlated with greater accuracy. Together these results suggest that a short course which educates people about how LLMs work and gives them practice in identifying AI-generated content could lead to higher discriminative accuracy.

Beyond recognizing AI-generated content, courses could be designed to aid people in reasoning critically about and responding to potentially deceptive or misleading arguments. A large body of work focuses on interventions that improve critical thinking skills in students \citep{abrami_strategies_2015}. Other studies have found that educational interventions can improve peoples' evaluation of informal arguments \citep{larson_improving_2009}, help people to identify false news stories \citep{guess_digital_2020}, and neutralise the effects of misinformation \citep{cook_neutralizing_2017}. Similar interventions could be developed to target AI-generated deceptive content specifically. 

Even if effective courses can be designed, however, a central challenge for this mitigation strategy is disseminating content and incentivising people to engage with it. Around 20\% of people in the United States lack moderate proficiency in even basic skills such as literacy, underlining the challenges of reaching a broad audience with more complex interventions such as critical thinking \citep{oecd_oecd_2013}.

\subsection{Regulation and Auditing}
Most of the strategies discussed above face a shared challenge of how to incentivise or enforce compliance. Successful mitigation of persuasive capabilities will likely require placing restrictions on model release, access, and usage \citep{burtell_artificial_2023}. Even if the technical challenges of aligning models and detecting manipulative behaviour can be solved, how can labs be incentivized to implement these strategies rather than racing ahead to deploy more capable models, and how can bad actors be prevented from training misaligned or otherwise harmful models?

A variety of types of regulation and legislation have been proposed to address these challenges \citep{janjeva_strengthening_2023, maslej_artificial_2024}. One major category of regulation places restrictions on model training and deployment to ensure that models meet certain safety standards \citep{shevlane_model_2023}. One prominent example was the vetoed California Senate bill SB1047, which would have required companies that conduct training runs using in excess of \$100m in compute to take various safety precautions including designing a safety plan and retaining a third-party auditor to ensure compliance. These types of regulations have the advantage that they do not inconvenience smaller and less well-resourced developers, and that deployment restrictions are flexible and could be changed as our understanding of capabilities and risks changes.

An alternative strategy is to mitigate harms after deployment. One approach is to require companies to include digital ``watermarks'' in AI-generated content \citep{dathathri_scalable_2024, park_ai_2024}, a requirement that also appears in the proposed California bill AB3211. This could allow other tools or users to identify when they are reading AI-generated text, which could mitigate some types of persuasion risks, including impersonation, fraud, and astroturfing. Other kinds of post-deployment approaches include adapting tort law to place the onus on developers for harms that befall users \citep{weil_tort_2024, willis_deception_2020}.

Any kind of regulation of AI faces myriad challenges, including defining harms, enforcement, and international cooperation. Defining potential harms from AI is particularly challenging as the technology is developing quickly. A wide scope will potentially stifle innovation, while a narrower scope risks allowing companies to engage in regulatory arbitrage---finding ways to technically comply with rules while undermining their intent. Historical examples include banks using securitization to circumvent capital requirements \cite{admati2013}, automobile manufacturers engineering software specifically to pass emissions tests \cite{coglianese2017}, and corporations implementing superficial box-ticking exercises to meet governance requirements \cite{romano2005}. Unless they are carefully written, AI regulations risk being susceptible to similar superficial compliance solutions. Secondly, even if laws around AI are well-written, they could be challenging to enforce because the hardware and algorithms used for these purposes are widely available and have many legitimate uses. At a broader scale, successful domestic enforcement will potentially do little good without international collaboration. While some proposals suggest that tracking of specialised hardware could help to ensure international enforcement \citep{shavit_what_2023}, this approach could become less viable if algorithmic improvements mean that more performant models can be trained on easily accessible hardware.

\section{Open Questions for Future Work}
A variety of open questions remain around how persuasive AI will develop. Here we collect those that we believe would most immediately advance understanding of the risk that persuasive AI systems pose and whether these risks can be mitigated. We organize promising avenues for future research into 5 overarching questions: how persuasive could AI systems be? How do AI systems persuade? What kinds of broader social impacts could AI persuasion have? Does the truth enjoy an advantage over falsehoods? And how effective are proposed mitigations?

\subsection{How persuasive could AI systems be?}
Existing work reviewed above suggests that AI systems can already create output that is roughly as persuasive as human writing across a wide range of domains and significantly more persuasive than the writing of even human experts (e.g. political consultants or the CDC) in others. Nevertheless, effect sizes tend to be small (around a 4-8\% change in self-reported beliefs). While these effects are typical or even relatively large for psychological persuasion studies, they are unlikely to be responsible for the most consequential harms outlined in theoretical models of risks. How much more persuasive are models likely to become in the next decades?

\citet{hackenburg_evidence_2024}'s scaling analysis suggests that, for static messages, gains in persuasiveness from scale alone could be small (less than a 1 percentage point increase for every order of magnitude increase in parameter count). Between 2017 and 2023, the parameter count of the largest LLMs increased roughly 5 orders of magnitude (from 65m parameters in the original transformers paper to unconfirmed reports of over 1 trillion in GPT-4), implying a trend slightly slower than one order of magnitude per year. If both of these trends were to hold, we would still not expect to see an increase of 10 percentage points (to $\sim$20\%) within a decade. In practice, many other factors could influence these trends. Scaling laws are often calculated to incorporate the total compute used to train a model rather than solely the number of parameters \citep{kaplan_scaling_2020}. Increases in the volume of data used to train models, as well as algorithmic improvements that allow compute to be used more efficiently \citep{erdil2022algorithmic}, could lead to more rapid increases in effective compute. However, many factors may conspire to slow scaling relative to the last 6 years, including practical challenges of managing large computing clusters, energy costs, and data availability. This analysis also raises the question of what counts as a large persuasive effect; the impact of models that can could people's consumption or voting behaviour by 15-20\% could be significant.

Moreover, few studies make effective use of the wide variety of mechanisms that are thought to allow LLMs to be so persuasive in theory. Interactive studies tend to show the largest effects, but as far as we are aware, no studies have explicitly measured how much more persuasive multi-turn interactions with LLMs are versus static messages. A small number of studies have tried prompting models in different ways \citep{durmus_measuring_2024, nisbett_how_2023}. \citet{jones_people_2024} evaluated more than 40 prompts for GPT-4 on the Turing test and found that performance varied drastically by prompt (from 6\% to 50\% pass rates). This suggests that prompt engineering could lead to significant increases in persuasive outcomes on other tasks. While some studies have shown that personalising messages to the user offers limited persuasive advantage \citep{hackenburg_evidence_2024, hackenburg_evaluating_2024}, others do show improvements \citep{matz_potential_2024, salvi_conversational_2024, simchon_persuasive_2024}, suggesting more ways in which persuasive effects could increase. Similarly, a few studies have suggested that the presentation of the identity of the model (e.g. as an AI, a person, or an expert) can influence the persuasiveness of the content it produces \citep{karinshak_working_2023, spitale_ai_2023}. To our knowledge, no studies have measured the impact on persuasive outcomes of chain-of-thought reasoning, fine-tuning, reinforcement learning for persuasiveness, access to tools such as browsing, or inference-time processes such as selecting among various candidate generations \citep{sumers_cognitive_2023}. Given the effectiveness of these techniques in other domains \citep{wei_chain--thought_2023}, it seems possible that they would also lead to greater persuasive effects.

Furthermore, many psychological theories of persuasion foreground the importance of social factors—such as trust and rapport—which have only begun to be explored in this literature. Factors such as likeability, reciprocity, authority, and social conformity have all been implicated in theories of belief change \citep{cialdini_influence_2003, henrich_evolution_2003}. While \citet{phuong_evaluating_2024} find that larger models elicit higher ratings on factors such as likeability, to our knowledge no studies have measured the impact of social factors on LLM persuasion outcomes (for example, by building a user's trust in a model before a critical persuasion event). Similarly, many studies point to the importance of affect, multimodality, and embodiment in attributing mindedness to artificial agents more broadly, which could in turn lead to larger effects of their output \citep{guingrich_chatbots_2024, scott_you_2023}. Already, users are interacting with agents that produce content via emotion-laden voices (e.g. Hume's EVI, and OpenAI's GPT-4o), with humanoid avatars (for example, Replika), and physical bodies \citep{driess_palm-e_2023}. It is vital to understand what impact these developments might have for models' ability to influence and manipulate people.

Future work should address these questions by testing the impact of these various mechanisms (individually and in concert) on persuasive outcomes. This work is potentially ethically fraught. Social scientists and safety researchers could inadvertently accelerate development of the very risks they hope to mitigate. Publication of methods and results should be accordingly thoughtful—balancing transparency with assessed risks of acceleration. Nevertheless, given the immense financial incentives of well-resourced groups such as advertisers, technology companies, and governments to achieve persuasive effects, on balance it seems likely that researchers can do more good than harm by highlighting risks before they are realised in order to motivate and test mitigations (see \citet{cotra_survey_2023} for discussion of acceleration risks of dangerous capability evaluations).

\subsection{How do LLMs achieve persuasive effects?}
A second key question concerns how LLMs achieve influence over people's beliefs when they do. This question partly overlaps with the previous one: it's important to quantify the extent to which prompting, fine-tuning, and social factors increase persuasiveness. In addition, answering this question involves analysing transcripts of interactions, classifying persuasion as rational or manipulative, and analysing the argumentative or rhetorical devices that LLMs employ. Doing so could help to build on research in social psychology and argumentation theory about what kinds of arguments people find persuasive \citep{okeefe_persuasion_2002}. More urgently, it could help researchers to design and assess the effectiveness of different mitigations.

Initial work along these lines \citep{bai_artificial_2023, carrasco-farre_large_2024} suggests that LLMs employ different persuasive tactics than humans. \citet{carrasco-farre_large_2024} found LLM arguments to be less readable and more complex, while \citet{bai_artificial_2023} found them to be more logical and factual than human-written arguments. Future work should investigate why differences between human and LLM arguments emerge: whether from features of the training data, reinforcement learning feedback, or whether training processes have discovered more effective persuasion techniques than humans generally use (as they have done in other domains; \citep{shin_human_2021}). 

Perhaps the most crucial question in this vein is the extent to which models use (and rely on) deceptive or manipulative techniques in order to achieve persuasive effects. For instance, if models tend to use deceptive techniques when arguing for false claims—such as stating mistruths, fallacious reasoning, or overt emotional manipulation—this would bolster the case that mitigations targeting process-level harms could effectively reduce negative outcomes \citep{el-sayed_mechanism-based_2024}. Alternatively, if models achieve deception by artfully arranging true facts, selectively omitting unhelpful ones, and constructing plausibly sound arguments that exploit legitimate epistemic uncertainty, it could suggest that detecting and countering deception will be more challenging. Future work should address these questions using argument mining and classification to analyse transcripts, as well as prompting models to use different techniques (e.g. to be more or less manipulative) to measure the impact this has on outcomes \citep{hagendorff_deception_2023}.

\subsection{What kinds of broader social impacts could persuasive LLMs have?}
The majority of extant studies focus on the effect of short interactions with LLMs on self-reported agreement with statements. By contrast, many of the potential risks of AI-generated persuasion concern societal outcomes such as influencing the ideology and behaviour of entire populations. It's unclear whether focused lab experiments will translate into society-wide effects due to noise, competition, attrition, saturation, and other countervailing forces. Future work should attempt to close this gap by exploring the effects of LLMs on broader societal outcomes \citep{weidinger_sociotechnical_2023}.

One crucial question is how robust the persuasive effects of LLMs are. When participants indicate changes in belief, do these last for months after the experiment or quickly decay? \citet{costello_durably_2024} followed up with participants 2 months post-intervention and found that the original 17\% difference between control and experimental groups had not significantly diminished. More measures like this could help to test whether other kinds of effects are this robust. Similarly, future work should investigate how generalisable effects are. If an intervention changes someone's belief about religion, does it also affect related beliefs (e.g. about evolution and morality)? Or are effects localised to the statements that conversations focus on?

Most studies focus on belief change. While this could be a precondition or a proxy for more impactful downstream effects, changing one's self-reported belief in an experiment is almost costless to participants and could be subject to demand characteristics. A handful of studies have tested whether LLM outputs can influence participant behaviour—for instance, donating to charity or clicking on a suspicious link \citep{phuong_evaluating_2024}. More studies that require participants to take actions with real-world consequences (or simulated consequences with real incentives) would help to better estimate the impact of interventions on how people will spend their time and money.

Real world conversations do not happen in a vacuum. People will likely be exposed to various viewpoints and people who interact with persuasive content will go on to interact with other people. Future work should investigate these second-order social effects of persuasion by presenting content in a more ecologically valid way, with a mixture of viewpoints. Transmission chain studies \citep{kirby_compression_2015, moussaid_amplification_2015} could be used to understand whether targets of AI-generated persuasion will themselves influence others. Simulations or experiments with small groups could provide insight into the marginal impact that LLM content will have in an already information-rich environment. Other important questions concern the longer term effects of exposure to persuasive AI content: do people lose trust or become saturated with information leading to a kind of ideological lock-in \citep{barnes_risks_2021}? Are LLMs effective at gathering information about individuals in a population and simulating their responses to hone persuasive strategies \citep{agnew2024illusion, floridi_hypersuasion_2024}? Finally, how are LLMs really being used in the real world? Are they already being used for persuasive functions on social media and in customer service \citep{shin_large_2023}? What impact is this having and how are people's attitudes toward AI changing?

\subsection{Will truth out?}
One of the most crucial questions in this domain is the extent to which the truth would enjoy an advantage over falsehoods in a world where LLMs can produce very persuasive content. This question presents itself at multiple scales. In a dyadic interaction, would a person be less likely to be persuaded of something false by an LLM than of the truth? In a debate between two LLMs, would the one arguing for the truth have an advantage? And at a societal scale, where many arguments are being made for a plurality of viewpoints, would the population's views tend toward the truth? 

If there is a strong advantage for the truth (and one that grows with the general persuasiveness of arguments) then risks from deception by LLMs might be greatly attenuated. In fact, more persuasive LLMs might raise general standards of discourse, as they find the most compelling arguments for unintuitive but sound positions. Alternatively, if this advantage is weak or nonexistent, then more capable models could lead to a kind of epistemic chaos, where persuasive arguments can be found for arbitrary positions.

There are theoretical reasons why there ought to be an advantage for the truth. True explanations are internally consistent, while defending some false positions will involve obfuscating or justifying internal inconsistencies. Similarly, true explanations are consistent with existing empirical observations, and make accurate predictions about the world. \citet{sperber_epistemic_2010} argue that our reasoning faculties evolved specifically to scrutinise potentially misleading claims, and so we might expect them to be well attuned to the kinds of falsehoods that LLMs could generate. Moreover, jury theorems suggest that aggregation of opinions leads to better decision-making that overcomes the bias and information-limits of individuals \citep{dietrich_jury_2023}. As our collective ability to gather data, process information, and scrutinise arguments increases, it seems plausible that our ability to identify and reject faulty arguments should improve. Some early evidence supports this optimistic position. \citet{fay_truth_2024} found that true messages (generated by both humans and LLMs) were rated as more likely to be true and more likely to be shared with others than false ones. However these conclusions are based on self-report rather than actual sharing. Both \citet{khan_debating_2024} and \citet{michael_debate_2023} found advantages for the truth in consultancy and debate experiments. However, their setup created strict advantages for the truth and it remains to be seen how widespread this advantage is for other types of questions.

There are alternative reasons to believe that misleading arguments may be equally persuasive as true ones, or even more persuasive in some cases.
While true claims are heavily constrained by the nature of the world, misleading arguments are free to take any form that suits a persuader (including starting from false premises, or using invalid reasoning to reach false conclusions from true premises). There are many factors which might influence belief acquisition beyond veracity, including the intuitive plausibility of the argument, the psychological needs of the persuadee, and apparent social consensus. Deceptive persuaders can optimise their arguments against these criteria without being constrained by the truth. Young-earth creationism, for instance, might continue to enjoy popularity because it is intuitively more plausible than evolution by natural selection, it fulfils psychological needs for meaning and purpose, and because (having reached a critical level of adoption in some communities) it is socially costly for individuals to abandon this view. In the same way that human society has developed ``supernormal stimuli''—such as sugar and pornography—which satiate historically adaptive cravings without providing attendant fitness benefits \citep{barrett_supernormal_2010, baudrillard_simulacra_1994}, a sufficiently good search process might be able to produce ``supernormal arguments'' which optimise responses from people's plausibility-seeking mechanisms without hewing closely to the truth.

There are many challenges to addressing this question. Studies like \citet{khan_debating_2024}'s are very valuable and should be extended to a wider variety of topics and conditions (including situations with multiple viewpoints, access to the internet, and real-world questions). One central challenge—which looms over much of this discussion—is how to define truth in the first place. Many important claims are disputed across the global population or within the scientific community. Other claims which used to enjoy global and scientific consensus are now widely disbelieved. One way of tackling this problem—that ``the truth'' is a moving target—is to ask instead whether people can be persuaded toward either side of an issue. Leaving aside the question of whether 2 + 2 \textit{truly} equals 4, if a conversation with an LLM could persuade a person either that it does or that it doesn't, then many of the risks highlighted above become more concretely likely.
That is, even if the truth value of a claim cannot be robustly verified for some domains, if LLMs are able to persuade people either for or against the same claim, their persuasive potential must not be sensitive to the ground-truth validity of the claim.

\subsection{How effective are proposed mitigations?}
A broad variety of mitigations have been proposed against persuasive and deceptive systems, but there is little empirical evidence to judge their effectiveness. Future work should test the individual claims of these strategies to understand whether they are workable in practice and to compare their effectiveness in reducing downstream harms.

Several strategies depend on being able to detect deceptive or manipulative content—either for monitoring systems, or as part of training models to avoid these behaviours. While there has been some work on detecting falsehoods \citep{boumber_domain-agnostic_2024}, more work is needed to understand how effective such systems are. In particular, if a person is interacting with a system in some context (e.g. as an assistant with access to the user's file system), is it possible for a monitoring system to tag mistruths purely from the transcript? Or would the monitoring system also need access to the user's system to run independent checks about the assistant's assertions? Similarly, are systems able to mislead users without producing outright lies? Iterative and adversarial evaluations, where a ``deceptive'' system is incentivized to find more subtle ways of misleading a user while evading a monitoring system, could be valuable in addressing these questions \citep{greenblatt_ai_2024}.

The promise of systems that preserve and empower users' autonomy is theoretically compelling. However, there has been little work on how to train such a system and how well it would work. Future work should investigate this question by measuring users' authentic endorsement of their own decisions at different time intervals after interacting with models. Other potential mitigations, such as providing users with training or tools such as a second model to debate with or access to the internet, should also be evaluated empirically to better understand how risks might translate into more realistic scenarios.

\section{Conclusion}
In theory, LLMs have the potential to produce incredibly persuasive content and change the epistemic landscape of human society. In practice, there is already widespread evidence that these systems produce content that is roughly as persuasive as human-written arguments. Current effects of persuasion are small, however, and it is unclear whether advances in model capabilities and deployment strategies will lead to large increases in effects or an imminent plateau. A variety of mitigations have been proposed to address potential risks, but there is limited empirical evidence about how well they work. Future work should investigate the broader social impacts that motivated actors could bring about using LLMs to persuade others, and the extent to which different responses could mitigate these risks.

\section*{Acknowledgements}

We would like to thank Open Philanthropy for providing funding for this work, and the following people for insightful comments on various stages of this manuscript: Micah Carrol, Josh Goldstein, Kobi Hackenberg, Samuel Taylor, Ned Cooper, Rasmus Overmark, Beba Cibralic, Jared Moore,
Winnie Street, Francis Rhys Ward, Hugo Mercier, Joel Lehman,  Maurice Jakesch, Beth Barnes, Tom Costello, Francesco Salvi, Ştefan Sarkadi, Max Nadeau,
Oisín Parkinson-Coombs, Nadja Winning, Sean Trott, and Pamela Riviere.

\bibliography{persuasion_review, custom}

\end{document}